%% file: main.tex
\documentclass[final]{cvpr}

\usepackage[font=small,labelfont=bf]{caption}
\usepackage[utf8]{inputenc} %
\usepackage[T1]{fontenc}    %
\usepackage{url}            %

\usepackage{times}
\usepackage{epsfig}
\usepackage{graphicx}
\usepackage{amsmath}
\usepackage{amssymb}
\usepackage{booktabs}       %
\usepackage{amsfonts}       %
\usepackage{xcolor}
\usepackage{nicefrac}       %
\usepackage{microtype}      %
\usepackage{graphicx}
\usepackage{amsmath}
\usepackage{overpic}
\usepackage{multirow}

\usepackage{bbm}

\newcommand{\modelname}{CoAM\xspace}
\newcommand{\networkname}{CD-UNet\xspace}

\usepackage{subfigure}

\newcommand{\eqnref}[1]{(\ref{#1})}

\usepackage[pagebackref=true,breaklinks=true,colorlinks,bookmarks=false]{hyperref}

\title{Co-Attention for Conditioned Image Matching}

\def\cvprPaperID{5368} %
\def\confYear{CVPR 2021}

\begin{document}

\author{Olivia Wiles\\
VGG, Dept.\ of Eng.\ Science\\
University of Oxford\\
{\tt\small ow@robots.ox.ac.uk}
\and
S\'{e}bastien Ehrhardt\\
VGG, Dept.\ of Eng.\ Science\\
University of Oxford\\
{\tt\small hyenal@robots.ox.ac.uk}
\and
Andrew Zisserman\\
VGG, Dept.\ of Eng.\ Science\\
University of Oxford\\
{\tt\small az@robots.ox.ac.uk}

}

\maketitle

\begin{abstract}
  \input{abstract}
\end{abstract}
\input{teaser}
\input{intro}

\input{relatedwork}

\input{method}

\input{experiments_stylization}

\input{conclusion}

\appendix

\input{appendix}

\clearpage 

{\small

\bibliographystyle{ieee_fullname}

\bibliography{shortstrings,egbib,vgg_local,vgg_other}
}

\end{document}

%% file: abstract.tex
We propose a new approach to determine correspondences between image
pairs  in the wild under large changes in illumination, viewpoint, context, and
material.  While other approaches find correspondences between pairs
of images by treating the images {\em independently}, we instead condition on {\em both} images to implicitly take
account of the differences between them.
To achieve this, we introduce (i) a spatial attention mechanism (a co-attention module,  \modelname) for conditioning the learned features on both images, and (ii)
 a distinctiveness score used to choose the best matches at test time.
\modelname~can be added to standard architectures and trained using self-supervision or supervised data, and 
achieves a significant performance improvement under hard conditions, e.g.~large viewpoint changes.
We demonstrate that models using \modelname achieve state of the art  or competitive results on a wide range of tasks: local matching, camera localization, 3D reconstruction, and image stylization.

%% file: teaser.tex
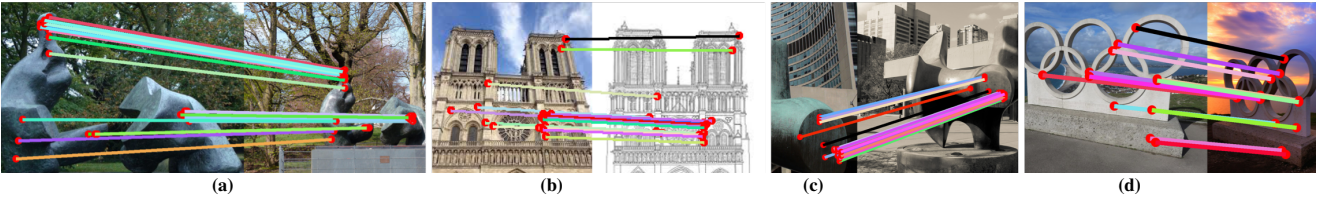
\begin{figure*}[h]
    \centering
    \begin{overpic}[width=\linewidth]{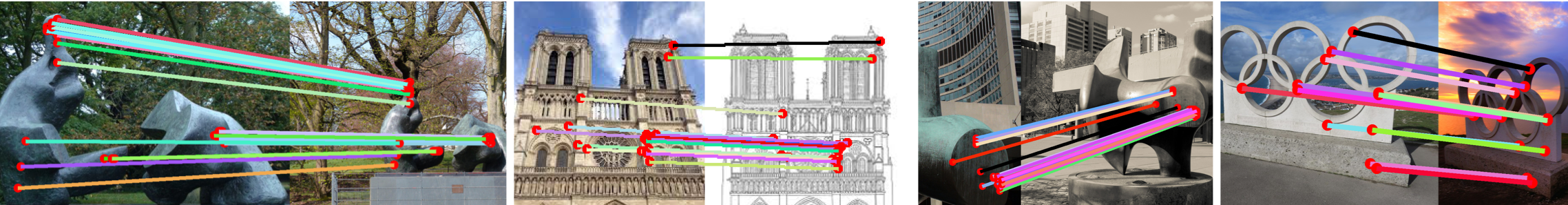}
    \put(16,-1.4){\bf \scriptsize (a)}
    \put(41,-1.4){\bf \scriptsize (b)}
    \put(61,-1.4){\bf \scriptsize (c)}
    \put(85,-1.4){\bf \scriptsize (d)}
    \end{overpic}
    \caption{{\bf Correspondences obtained with the \modelname~model, which is augmented with an attention mechanism. } These demonstrate the model's robustness in the face of challenging {\em scene shift}: changes in illumination (a,d), viewpoint (a-d), context (a,d), or style (b). }
    \label{fig:my_label}
    \vspace{-0.75em}
\end{figure*}

%% file: intro.tex
\vspace{-1em}

\section{Introduction}
\label{sec:intro}

Determining correspondence between two images of the same scene or object is a fundamental challenge of computer vision, important for many applications ranging from optical flow and image manipulation, to 3D reconstruction and camera localization.
This task is challenging due to {\em scene-shift}: two images of the same scene can 
differ dramatically due to variations in illumination (e.g.~day to night), viewpoint, texture, and season (e.g.~snow in winter versus flowering trees in spring).

Methods that solve the correspondence task typically follow a {\em detect-and-describe}  approach:
first they {\em detect} distinctive regions~\cite{Bay06,Harris88,Lowe04,Mikolajczyk05,Rublee11} and then {\em describe} these regions using
descriptors~\cite{Bay06,Calonder10,Ke04,Leutenegger11,Lowe04,Rublee11} with varying degrees of invariance to scale, illumination, rotation, and affine transformations. These descriptors are then matched between images by comparing descriptors exhaustively, often using additional geometric constraints~\cite{Hartley00}. Recent approaches have sought to learn either or both of these components~\cite{Balntas16,Choy16,Detone18,Dusmanu19,Germain20,Li20,Ono18,Rocco20,Rocco18b,Savinov17,Simo15,Simonyan14,Yi16,Zhang18Feature}.
These methods typically only find matches at textured locations, and do not find matches over smooth regions of an object.
Additionally, finding these repeatable detections with invariance to scene-shift is challenging~\cite{Balntas17,Sattler18,Schoenberger17}.

If prior knowledge is assumed, in terms of limited camera or temporal
change (as in optical flow computation in videos), then a {\em dense-to-dense} approach can be used for
pairs that have limited scene shift.
In this case, methods typically obtain a dense feature map which is compared from one image to another
by restricting the correspondence search to a small support region in the other image (based on the prior knowledge).
Spatial and smoothness constraints can
additionally be imposed to improve results~\cite{Choy16,Fathy18,Liu10b,Savinov17Features,Tola08,Wang19b}.

We focus on the cases where there is  potentially significant scene shift (and no prior knowledge is available), and 
introduce a new approach for obtaining correspondences between a {\em pair} of images.
Previous methods learn descriptors for each image {\em without} knowledge of the other image.
Thus, their descriptors must be invariant to changes -- e.g.\ to scale and illumination changes. 
However, as descriptors become increasingly invariant, they become increasingly ambiguous to match (e.g.~a constant descriptor is invariant to everything but also confused for everything).
We forsake this invariance and instead condition the descriptors on {\em both} images.
This allows the descriptors to be modified based on the differences between the images (e.g.~a change in global illumination).
Traditionally, this was infeasible, but we can learn such a model efficiently using neural networks.

To achieve this we introduce a network ({\networkname}), which consists of two important components.
First, a new spatial {\em {\bf C}o-Attention Module} (\modelname) that can be `plugged into' 
a UNet, or similar architectures developed for single image descriptors, in order to generate
descriptors conditioned on the pair of images.
Second, we introduce  a {\em {\bf D}istinctiveness} score in order to select the best matches from these descriptors.

We further investigate the utility of the  \modelname~under both supervised and self-supervised training.
In the latter case, we augment the recent self-supervised approach of learning camera pose of~\cite{Wang20} by using 
\modelname{}s~in a plug-and-play fashion.
We evaluate these trained  models on a variety of tasks:
local matching, camera localization, 3D reconstruction, and style transfer.
We improve over state-of-the-art (sota) models, especially under challenging conditions,  and achieve sota or comparable on all tasks.

In summary, we present a key  insight: that {\bf conditioning} learned descriptors on both images should allow for improved correspondence matching under challenging conditions.
As will be seen, \networkname~is simple and scalable and eschews
a number of techniques used by other methods to improve matching  
performance: high dimensional descriptors (we use a 64D descriptor, half the size of the current smallest descriptor),  and multiple scales (we only operate at a single scale, whereas other methods use multiple scales).

%% file: relatedwork.tex
\section{Related Work}

In this section, we review related work on finding correspondences beyond the local descriptors discussed in \secref{sec:intro}.
As there is a large amount of relevant research, we focus on the most relevant work in each category.

\noindent {\bf Correspondences using an attention mechanism.}
Our architecture can be viewed as a generalization of the standard correlation layer used in training end-to-end models for optical flow \cite{Dosovitskiy15,Ilg17,Sun18b}, stereo \cite{Kendall17} or correspondence estimation \cite{Fathy18,Lai20,Savinov17Features,Vondrick18,Wang20,Wang19b}. 
This correlation layer (or attention mechanism) is used to compute a cost volume of matches from the learned descriptors.

In correspondence estimation, the learned descriptors are limited in spatial resolution \cite{Lai20,Vondrick18,Wang19b} so that the entire volume can be computed. 
This is too coarse for geometric matching, so other methods use a hierarchical approach \cite{Fathy18,Savinov17Features,Wang20}.
In optical flow \cite{Dosovitskiy15,Ilg17,Ranjan17,Sun18b} and stereo \cite{Kendall17}, the cost volume is only applied within a limited support region for a single descriptor (e.g.~a square region  or a raster line) and typically at a lower resolution.
Moreover, these methods implicitly assume photometric consistency between frames: their quality degrades the more the frames differ in time, as the pose and viewpoint progressively change.

Unlike these methods, we apply attention at multiple stages in our network, so that the final descriptors {\em themselves} are conditioned on both images. 
This should be beneficial for challenging image pairs where one, final comparison is unable to encompass all possible scene-shifts between two images.
To find matches at the image level without performing an exhaustive comparison, we use a modified hinge loss to enforce that true descriptors are nearby and false ones further away.

\noindent {\bf Dense correspondence matching with prior knowledge.} 
Given a static scene and initial camera estimation, a second algorithm, e.g.~PatchMatch \cite{Barnes09,Schoenberger16b}, can be used to find dense correspondences between the images and obtain a full 3D reconstruction. 
If the images have been rectified using multiple-view geometry \cite{Hartley00} and have limited {\em scene shift}, stereo algorithms such as \cite{Hanna74,Pollefeys99,Sun05,Woodford09a} (and reviewed by \cite{Szeliski10}) can be used to obtain a full 3D reconstruction.

While not directly related, \cite{Kim18,Shen20} condition on a second image by iterative warping.
This requires multiple passes through a network for each image pair and uses pre-trained descriptors as opposed to training end-to-end.

Also related are approaches that seek to learn correspondence between similar scenes \cite{Liu10b} or instances of the same semantic class \cite{Kim18,Novotny17A,Rocco17,Rocco18a}.

\noindent {\bf Local descriptors for image retrieval.} Another form of correspondence is to find relevant images in a database using a query image. Related works use an aggregation of local descriptors from a CNN  \cite{Cao20,Tolias20}.
Again, these methods generate descriptors for the dataset images independently of the query image, whereas the descriptors we extract for the input image are conditioned on both images.

\noindent {\bf Stylization for robust correspondence matching.} Our idea of conditioning the output of one image on another has interesting connections to stylization and associated generative models \cite{Karras19,Park19,Ulyanov16}.
Additionally, a recent line of work studies how training on stylized images can improve robustness in correspondence matching \cite{Melekhov20}.
As opposed to enforcing invariance to style, \networkname~and the other architectures considered, learn how to leverage differing styles (as the precise style may be useful) via our \modelname.

%% file: method.tex
\input{figuremethod}

\section{Method}
Our task is to find dense correspondences between a pair of images of the same scene.
This proceeds in two stages. 
The first stage obtains dense descriptor vectors for each image and a distinctiveness score.
The descriptors are conditioned on {\em both} images so they only have to be invariant to the changes particular to that pair of images.
The second stage compares these descriptor vectors to obtain a set of high quality matches.
We first describe in \secref{sec:attnetarchitecture} our full architecture \networkname, and how it is trained in \secref{sec:suptrainingandlossfunctions}.
\networkname consists of a set of Co-Attention Modules (\modelname{}s) and a distinctiveness score regressor, which are incorporated into a
standard UNet architecture. Then, in \secref{sec:capsarchitecture}, we
describe how \modelname is incorporated into the recent CAPSNet architecture~\cite{Wang20} and trained
in a self-supervised manner.

\subsection{A UNet encoder-decoder with \modelname}
\label{sec:attnetarchitecture}
The architecture for obtaining descriptor vectors and a distinctiveness score for one image $I^1$ (\figref{fig:overviewpipeline}),  is composed of four components.
The first component, the {\bf encoder}, projects both images $I^1, I^2$ to obtain feature maps at two resolutions: $f^i_L$, $f^i_S$.
The second component, the {\bf attention mechanism (\modelname)}, is used to determine spatial correspondence between the feature maps of the different images and obtain conditioned feature maps.
The third component, the {\bf decoder}, concatenates the conditioned feature maps with the original feature maps. These are decoded to obtain a grid of spatial descriptor vectors $D^1$ (which are conditioned on both images).
The final component, the {\bf regressor}, learns a distinctiveness score for each grid position, which encodes how likely the match is to be accurate.
To obtain descriptor vectors $D^2$ for the other image, we operate precisely as described above, except that the order of the input images is flipped.
This gives a grid of descriptor vectors $D^1$, $D^2$ for images $I^1$, $I^2$ respectively.

\noindent {\bf Encoder.}
Given two images of the same scene, $I^1 \in \mathbbm{R}^{H \times W \times 3}$ and $I^2 \in \mathbbm{R}^{H \times W \times 3}$, we obtain spatial feature maps: $f^i_L$ and $f^i_S$ at a $l$arger and $s$maller resolution.
These will be concatenated within a UNet framework \cite{Ronneberger15} and injected into the decoder.
A CNN with shared parameters is used to encode the images and obtain these spatial feature maps.
In practice, we use the feature maps after the last two blocks in a ResNet50 \cite{He15} architecture.

\noindent {\bf \modelname~Attention Module.} 
We wish to concatenate features from both images in order to condition the model on both input images.
However, for a given spatial location, the relevant (corresponding) feature in the other image may not be at the same spatial location.
As a result, we use an attention mechanism to model long range dependencies.

In detail, the attention mechanism is used to determine where a location $i$ in one set of features $g$ from one image should attend to in another set of features $h$ from another image \cite{Vondrick18}.
For each location $i$ in $g$, it obtains a feature $\hat{g_i}$ that is a weighted sum over all spatial features in $h$ where $A$ is the similarity matrix comparing $g$ and $h$ using the inner product followed by the softmax normalization step.
\begin{equation}
\label{eq:attention}
    \hat{g}_i = \sum_{j} A_{ij} h_j \quad \quad \quad A_{ij} =  \frac{\texttt{exp}(g^T_i h_j)}{\sum_k \texttt{exp}(g^T_i h_k)}
\end{equation}

To apply this attention mechanism, we operate as follows for $f^1_L$ (and similarly for $f^1_S$).
First,   to perform dimensionality reduction (as is standard), the features are projected with two MLPs $g^1(\cdot), g^2(\cdot)$: $g = g^1(f^1_L)$, $h = g^2(f^2_L)$.
The attended features $\hat{f}^1_L$ are then computed using the projected features as in \eqnref{eq:attention}.
This gives a new feature map of the features in $I^2$ at the corresponding position in $I^1$.

\noindent {\bf Decoder: Conditioned Features.}
The attended features are incorporated into a UNet \cite{Ronneberger15} architecture to obtain a grid of spatial descriptors $D^1 \in \mathbbm{R}^{H \times W \times D}$ (\figref{fig:overviewpipeline}).
The attended features $\hat{f}^1_L$ and $\hat{f}^1_S$ are concatenated with the original features and passed through the decoder portion of the UNet.
The resulting feature map is $L2$ normalized over the channel dimension to obtain the final descriptors.
This step ensures that the final descriptors are conditioned on both images.

\noindent {\bf Regressor: Distinctiveness Score.}
We regress a distinctiveness score $r(\cdot)_{ij} \in [0,1]$, for each pixel $(i,j)$, which approximates its matchability and is used at test time to select the best matches.
$r(\cdot)_{ij}$ approximates how often the descriptor at $(i,j)$ is confused with negatives in the other image. 
If it is near $1$, the descriptor is uniquely matched; if it is near $0$, the descriptor is often confused.
To regress these values, we use an MLP, $r(\cdot)$, on top of the unnormalized descriptor maps.

\noindent {\bf Determining Matches at Test Time.}
\label{sec:correspondencetest}
We want matches at locations $k$ and $l$ in images $I^1$ and $I^2$ respectively that are accurate and distinctive (e.g.~no matches in the sky).
We use the scalar product to compare the normalized descriptor vectors to find the best matches and the distinctiveness score to determine the most distinctive matches.
The following similarity score $c_{kl}$ combines these properties  such that a value near $1$ indicates a distinct and accurate match:
\begin{equation}
\label{eq:similarity}
    c_{kl} = r(D^1_k) r(D^2_l) \left[ \left( D^1_{k} \right)^T D^2_{l} \right] .
\end{equation}

Finally, we select the best $K$ matches.
First, we exhaustively compare all descriptors in both images. Then, we only select those matches that are mutual nearest neighbours: e.g.~if the best match for location $m$ in one image is location $n$ in another, and the best match for location $n$ is  $m$, then $(n, m)$ is a good match. So if the following holds:
\begin{equation}
    m = \texttt{argmax}_j c_{nj} \quad \text{and} \quad n = \texttt{argmax}_i c_{im}.
\end{equation}
These matches are ranked according to their similarity score and the top $K$ selected.

\subsection{Supervised Training and Loss Functions}
\label{sec:suptrainingandlossfunctions}

\noindent {\bf Selecting Correspondences at Train Time.}
\label{sec:expsampling}
Given a ground-truth correspondence map, we randomly select $L$ positive correspondences.
For each positive correspondence, we randomly select a large number ($N=512$) of negative correspondences.
These randomly chosen positive and negative correspondences are used to compute both the distinctiveness and correspondence losses.

\noindent {\bf Correspondence Loss.}
The correspondence loss is used to enforce that the normalized descriptor maps $D^1$ and $D^2$ can be compared using the scalar product to obtain the best matches.
At a location $i$ in $D^1$ and $j$ in $D^2$ then the standard Euclidean distance metric $d(D^1_i, D^2_j)$ should be near $0$ if the corresponding normalized descriptor vectors are a match.

To train these descriptors, we use a standard contrastive hinge loss to separate true and false correspondences (we consider other contrastive losses in the appendix).
For the set $\mathcal{P}$ of $L$ true pairs, the loss $\mathcal{L}_{p}$ enforces that the distance between descriptors is near $0$. 
For the set $\mathcal{N}$ of $LN$ negative pairs, the loss $\mathcal{L}_n$ enforces that the distance between descriptors should be above a margin $M$.
\begin{align}
    \mathcal{L}_{p} &= \frac{1}{L} \sum_{(x,y) \in \mathcal{P}} d(D^1_{x}, D^2_{y})  \\
    \mathcal{L}_{n} &= \frac{1}{LN} \sum_{(x, \hat{y}) \in \mathcal{N}} \max (0, M + c_{x}- d(D^1_{x}, D^2_{\hat{y}})).
\end{align}

$c_{x} = d(D^1_{x}, D^2_{y}), (x,y) \in \mathcal{P}$ re-weights the distance of the false correspondence according to that of the positive one: the less confident the true match, the further the negative one must be from $M$~\cite{Dusmanu19}.

\input{hpatches_results}

\noindent {\bf Distinctiveness Loss.}
To learn the $r(\cdot)$ MLP, we need an estimate of how often a descriptor in one image is confused with the wrong descriptors in the other image.
Given a set $\mathcal{N}$ of $N$ negative matches in the other image and the margin $M$, the number of times a descriptor at location $x$ is confused is $m_{x} = \sum_{\hat{y} \in \mathcal{N}} \mathbbm{1}(d(D^1_{x}, D^2_{\hat{y}}) < M)$. 
This value is used to regress $r(\cdot)$, which is near $1$ if the feature has a unique match  (the true match), near $0$ otherwise ($\tau$ is a hyper-parameter set to $\frac{1}{4}$):
\begin{equation}
    \mathcal{L}_r = \frac{1}{L} \sum_{(x,\cdot) \in \mathcal{P}} | r(D^1_{x}), \frac{1}{(1 + m_{x})^\tau} |_1.
\end{equation}

\noindent {\bf Training Setup.}
\networkname~is trained on MegaDepth \cite{Li18}, which consists of a variety of landmarks, registered using SfM \cite{Schoenberger16a}.
As each landmark consists of many images taken under differing conditions, we can obtain matches between images that are unmatchable when considered independently.

We train the features end-to-end, but train the distinctiveness score separately by not allowing gradients to flow.
In practice we backpropagate on all randomly chosen positive pairs $\mathcal{L}_p$,  negative pairs $\mathcal{L}_n$, and additionally the hardest $H=3$ negative pairs for each positive pair. 

The model is trained with a learning rate of $0.0001$, the ADAM optimizer \cite{Kingma14}, a batch size of $16$, $M$=$1$, $L$=$512$, and $N$=$512$.
At train time we use an image size of $256$, at test time an image size of $512$.
We use $K$=$2000$ for HPatches and Aachen,  and $K$=$8192$ when performing SfM. 
For SfM, we find it is important to use more, rougher correspondences to obtain more coverage in the 3D reconstruction.

\subsection{Self-supervised training -- the CAPSNet \cite{Wang20} with \modelname}
\label{sec:capsarchitecture}
In this section we describe how \modelname~can be added to the CAPSNet architecture of~\cite{Wang20} and
trained using the self-supervised framework of~\cite{Wang20}.

CAPSNet consists of a UNet style architecture, which predicts features at a coarse and fine level.
The matches at a coarse level are used to guide feature matching at the finer level.
These features are trained using two losses.
First, an epipolar loss enforces that matches should satisfy epipolar constraints.
Second, a cycle consistency loss enforces that, for a match between two images, the best match for the local descriptor in one image should also be the best match for the local descriptor in the other.
Using this approach, the authors achieve high quality results at pose estimation on the challenging MegaDepth test set.

As the descriptor model is a UNet style architecture,  and it is  trained in an end-to-end fashion, we operate in a very similar manner to the UNet architecture with \modelname~of \secref{sec:attnetarchitecture}, by again
adding \modelname{}s to condition descriptors on both images.
We use the \modelname to inject attended features from the other image at either a coarse level or at a fine and coarse level (precise details are given in the appendix).
In both cases, this leads to an addition of less than 15\% of the total weights of the original network.

The loss functions used to train the conditioned local descriptors are unchanged from the original CAPSNet work.

\noindent {\bf Training Setup.}
We train the model as done in \cite{Wang20}: for 200K iterations, using a batch size of $6$ images, and an image size of $480\times640$.

\subsection{Discussion}

Here we discuss some of the benefits of conditioning using \modelname~as opposed to operating directly on local descriptors and keypoints as done in SuperGLUE \cite{Sarlin20}.
First, our module is trained end-to-end and does not introduce an extra step in the matching pipeline of comparing pre-trained descriptors.
Second, our descriptors are learned, so our method is not dependent on the quality of the extracted descriptors.
Finally, SuperGLUE scales with the number of extracted keypoints, hampering its performance and utility on tasks that require finding a large number of correspondences (e.g.~SFM).
As the \modelname~is plugged in as a component of our network, our method scales with image size.
For reference, on a single GPU, to extract 2k keypoints on a $256\times 256$ image, our method runs in 97ms while SuperGLUE would add an overhead of $\approx$270ms as reported in the original paper. Further, our method would scale with little overhead to more keypoints at the given image size.

Our method requires an exhaustive match of all image pairs. While we find that we can run the full, exhaustive pipeline on reasonably large datasets ($\approx$ 1500 images) in \secref{sec:sfmexperiments}, we envision two stages when using our method in truly large scale settings.
First, a coarser, faster method can be used as a preprocessing step to remove spurious pairs and our method subsequently used in a second stage to find high quality correspondences.

%% file: figuremethod.tex
\begin{figure*}
    \centering
    \includegraphics[width=0.8\linewidth]{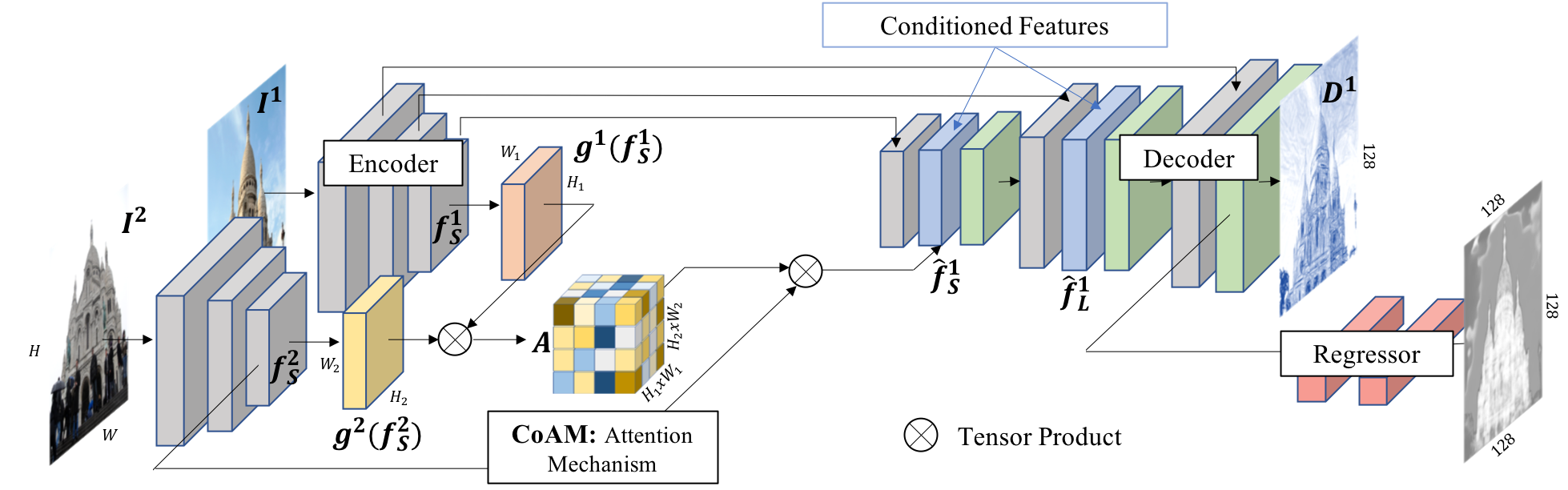}
    \caption{{\bf Overview of \networkname~for obtaining co-attended descriptors.}
    Descriptor vectors $D^1$ for one input image $I^1$ are conditioned on another $I^2$ using our   \modelname. This module can be applied at multiple layers in the model hierarchy (we show one for clarity). The conditioned features are then {\em decoded} to obtain $D^1$. 
    We also {\em regress} a distinctiveness mask which is used at test time to ignore unmatchable regions (e.g.~the sky or regions visible in only one image). The descriptor vectors $D^2$ for $I^2$ are obtained by swapping the input images. }
    \label{fig:overviewpipeline}
\end{figure*}

%% file: hpatches_results.tex
\begin{figure*}[t]
    \begin{minipage}{0.6\linewidth}
    \centering
    \includegraphics[width=\linewidth]{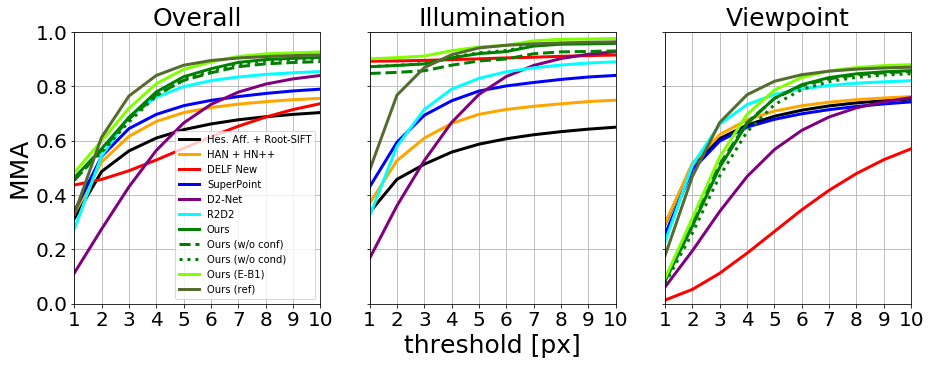}
    \end{minipage} \quad
    \begin{minipage}{0.4\linewidth}
    \centering
    \scriptsize
    \begin{tabular}{@{}lc@{}}
        \toprule
         Method & \# Matches \\
         \midrule
         Hes.~det.~+ RootSift \cite{Arandjelovic12,Lowe04} & 2.8K \\
         HAN + HN++ \cite{Mishkin18,Mishchuk17} & 2.0K \\
         LF-Net \cite{Ono18} & 0.2K \\
         SuperPoint \cite{Detone18} & 0.9K \\
         DELF \cite{Noh17} & 1.9K \\
         D2 Trained (SS) \cite{Dusmanu19} & 2.5K \\
         R2D2 \cite{Revaud19} & 1.8K \\
         \midrule
         Ours & 2.0K \\
         \bottomrule
    \end{tabular}
    \end{minipage}
    \caption{{\bf HPatches \cite{Balntas17}.} Comparison with sota using the mean matching accuracy for different pixel thresholds on the HPatches dataset. We also report the mean matches extracted per image pair. For this dataset, one desires more matches with high accuracy. 
    Our method achieves superior performance  when images vary by illumination for all thresholds,  and by viewpoint for thresholds $>6$px. By a simple refinement strategy ({\bf ours (ref)}), we achieve sota  for all thresholds on both viewpoint and illumination. }
    \label{fig:hpatchesresults}
\end{figure*}

%% file: experiments_stylization.tex
\section{Experiments I: Supervised Co-AM}
\label{sec:fullpipelineexperiments}
In this section we evaluate the \networkname architecture (UNet encoder-decoder with CoAM and distinctiveness score as in \figref{fig:overviewpipeline}) on
four  challenging downstream tasks  under full supervision.
In  \secref{sec:capsnetexperiments} the benefits of the co-attention module 
are evaluated under self-supervised training~\cite{Wang20}.

The first task directly assesses how well \networkname can estimate correspondences between images pairs.
The second  task uses the correspondences to perform camera localization.
In these tasks we ablate
the utility of the \modelname~and distinctiveness score components of the architecture.
The third task obtains high quality 3D reconstructions  in challenging
 situations, with a large amount of {\em scene shift}. 
The final task is stylization, and assesses \networkname's matches, when extracted in a dense manner, on a downstream task.

In general we find that \networkname achieves state of the art or comparable results and that the \modelname~is useful especially in challenging conditions (e.g.~when there is a large viewpoint change).

The appendix includes further ablations to validate our choices (e.g.~the loss function and grid size) and datasets (e.g.~(1) YFCC100M \cite{Thomee16} which shows our superior results and the utility of both the distinctiveness score and the \modelname, and (2) a new, challenging SFM dataset).
Finally, it includes qualitative samples for each of the experiments discussed in the following, including HPatches, Aachen, SFM, and stylization.

\noindent {\bf Ablations.}
The full model uses the ResNet50 \cite{He15} backbone, the \modelname{}s and the distinctivness score to reweight  matches.
We ablate multiple variants. 
The first ({\bf ours}) is our full model.
The second ({\bf ours w/o conf}) is our model without the distinctiveness score but only the scalar product.
The third ({\bf ours w/o cond}) is our model without conditioning (i.e.~the \modelname{}s).
The final variant ({\bf ours-E-B1}) is our full model but using an EfficientNet-B1 backbone \cite{Tan19}. This ablation uses a smaller (7M params vs 23M params) and faster (0.7GFlops vs 4.1GFlops) backbone architecture; it is more suitable for practical applications.

\subsection{Correspondence Evaluation}
We test our model on local matching by 
evaluating on the HPatches \cite{Balntas17} benchmark.
We compare to a number of baselines and achieve state-of-the-art results.

\noindent {\bf HPatches Benchmark.}
The HPatches benchmark evaluates the ability of a model to find accurate correspondences between pairs of images, related by a homography, that vary in terms of illumination or viewpoint.
We follow the standard setup used by D2Net \cite{Dusmanu19} by selecting 108 of the 116 sequences which show 6 images of larger and larger illumination and viewpoint changes.
The first image is matched against the other 5, giving 540 pairs.

\input{aachen_results}

\noindent {\bf Evaluation Setup.}  We
follow the evaluation setup of D2Net \cite{Dusmanu19}.  For each image
pair, we compute the number of correct matches (using the known
homography) and report the average number of correct matches as a
function of the pixel threshold error in \figref{fig:hpatchesresults}.
We then compare to a number of detect-then-describe baselines used in
D2Net using their software: RootSIFT \cite{Arandjelovic12,Lowe04} with
the Affine keypoint detector \cite{Mikolajczyk04b}, HesAffNet
\cite{Mishkin18} with HardNet++ \cite{Mishchuk17}, LF-Net
\cite{Ono18}, SuperPoint \cite{Detone18}, DELF \cite{Noh17}; as well as to  D2Net
\cite{Dusmanu19} and R2D2 \cite{Revaud19}.  These methods vary in
terms of whether the detector and descriptors are hand crafted or
learned.

\noindent {\bf Results.}
As shown in \figref{fig:hpatchesresults}, all variants of our model outperform previous methods for larger pixel thresholds, demonstrating the practicality and robustness of our approach.
In comparison to other methods, \networkname~performs extremely well when the images vary in illumination:  it outperforms all other methods. 
\networkname~is superior under viewpoint changes for larger pixel thresholds ($>6$px).
Using the smaller, more efficient ({\bf ours-E-B1}) actually improves performance over the larger ResNet model ({\bf ours}).
A simple refinement strategy (described in the appendix) boosts our model's performance under viewpoint changes, giving results superior or comparable to sota methods for all thresholds for viewpoint and illumination changes.
Compared to the other evaluation datasets, e.g.~\cite{Sattler18} below, 
the components of our model have a limited impact on performance on this benchmark, presumably because this dataset has less {\em scene
shift} than the others.

\subsection{Using Correspondences for 3D Reconstruction}
In this section, we evaluate the robustness of our approach on images that vary significantly in terms of illumination and viewpoint,  and our model's ability to scale to larger datasets. \networkname~achieves sota or comparable results on all datasets.

\subsubsection{Camera Localization}
{\bf Aachen Benchmark.}
In order to evaluate our approach under large illumination changes, we use the Aachen Day-Night dataset \cite{Sattler18, Sattler12}.
For each of the 98 query night-time images, the goal is to localize the image against a set of day-time images using predicted correspondences.

\noindent {\bf Evaluation Setup.} 
The evaluation measure is the percentage of night time cameras 
that are localized within a given error threshold~\cite{Sattler18}. 
We use the pipeline and evaluation server of~\cite{Sattler18} with the matches automatically obtained with our method (\secref{sec:correspondencetest}).
We compare against RootSIFT descriptors from DoG keypoints \cite{Lowe04}, HardNet++ with HesAffNet features \cite{Mishchuk17,Mishkin18}, DELF \cite{Noh17}, SuperPoint \cite{Detone18},  D2Net \cite{Dusmanu19} and DenseSFM \cite{Sattler18}. 

\noindent {\bf Results.}
\tabref{tab:aachenexperiments} shows that our method does comparably or better than other sota approaches.
They also show the utility of the distinctiveness score and \modelname.
These results imply that the traditional approach of first finding reliably detectable regions may be unnecessary; using a grid to exhaustively find matches is, perhaps surprisingly, superior in this case.
These results also show that our architectural improvements (i.e.~using an attention mechanism and distinctiveness score) boost performance and that an efficient architecture ({\bf Ours-E-B1}) has  a small impact on performance.

\input{sfmresults_stylization}

\input{stylisation_results_stylisation}

\subsubsection{Structure from Motion (SfM)}
\label{sec:sfmexperiments}
The objective here is to evaluate
the correspondences obtained with our model for the task of 3D reconstruction.

\noindent {\bf SfM Dataset.} 
The assessment is on the standard SfM Local Feature Evaluation Benchmark \cite{Schoenberger17} that
contains many ($\approx 1500$) internet images of three landmarks: Madrid Metropolis, Gendarmenmarkt, and Tower of London.

\noindent {\bf Baselines.}
We compare to SIFT \cite{Lowe04}.
This method first finds repeatably detectable regions for which features are extracted and compared between images.
This method works well when there are distinctive textured regions that can be found and matched.
Our method, however, conditions on both images, so our approach should be more robust when there are fewer textured regions or where there are significant variations between the images as it can make use of auxiliary information from the other image.
Additional, less challenging baselines are given in the appendix.

\noindent {\bf Results.}
 \tabref{tab:reconstructionresults} shows that, using \networkname, we consistently register more images and obtain more sparsely reconstructed 3D points (visualizations are in the appendix).  
However, the pixel error is higher and there are fewer dense points.
These differing results are somewhat explained by the implicit trade off between number of points and reprojection error \cite{Wang20}.
However, clearly our results are competitive with SIFT.

\input{attentioncoam_stylization}

\subsection{Using Correspondences for Stylization}
Previously, we focused on using our matching pipeline for extracting a set of correspondences to be used for localization and 3D reconstruction.
Here we evaluate how well our features can be used for a task that requires dense matching: stylization.
The goal is, given two images $I_s$, $I_v$ of the same scene, to generate an image with the style of $I_s$ but the pose and viewpoint of $I_v$.

\noindent {\bf Setup.}
To achieve this, we first use \networkname to transform $I_s$ into the position of $I_v$.
The approach is simple: instead of only choosing the mutual nearest neighbours as in \eqnref{eq:similarity}, we consider the best match for {\em every} pixel location.
Then, the color of the best match in $I_s$ is used to color the  corresponding location in $I_v$. 
This gives the {\em sampled image}.
The next step is to remove artefacts. We do this by training a refinement model on top of the sampled image in order to obtain an image $I_g$ in the pose of $I_v$ and style of $I_s$.
Full details of the architecture and training are given in the supp.

\noindent {\bf Relation to Pix2Pix \cite{Isola17}.}
In a standard image to image translation task (e.g.~Pix2Pix \cite{Isola17}), the two images (e.g.~image and semantic map) are aligned.
In our case, the images are not aligned. We effectively use our correspondences to align the images and then run a variant of Pix2Pix.

\noindent {\bf Experimental Setup.}
To evaluate our results, we use the test set of the MegaDepth dataset (these are landmarks unseen at training time).
We randomly select $400$ pairs of images and designate one the viewpoint $I_v$ image and the other the style image $I_s$.
We task the models to generate a new image $I_g$ with the style of $I_s$ in the viewpoint of $I_v$.
From the MegaDepth dataset, we can obtain ground truth correspondence for regions in both images and so the true values of $I_g$ for this region.
The reported error metric is the mean $L1$ distance between the generated image and true value within this region.

\noindent {\bf Results.} 
We compare against a stylization approach that uses semantics to perform style transfer \cite{Luan17} in \figref{fig:stylisationresults}.
We also determine the $L1$ error for both setups and obtain $0.22$ for \cite{Luan17} and $0.14$ for our method, demonstrating that our method is more accurate for regions that can be put in correspondence.
The qualitative results demonstrate that our method is more robust, as \cite{Luan17} produces poor results if the semantic prediction is wrong and sometimes copies style from $I_v$ as opposed to $I_s$ (e.g.~it creates a colored $I_g$ image when $I_s$ is grey-scale).
As we sample from $I_s$ in the first step and then refine the sampled image, our model rarely copies style from $I_v$.
Finally, our full method runs in seconds at test time whereas~\cite{Luan17} takes  minutes due to a computationally intensive iterative refinement strategy. 

\input{megadepth_results_ss}

\section{Experiments II: \modelname~with CAPSNet}
\label{sec:capsnetexperiments}
Next, we evaluate \modelname~when injected into the CAPSNet architecture \cite{Wang20} and trained in a self-supervised manner.
We again validate our hypothesis that conditioning on two images is preferable in this setting, as it improves results on the downstream task of pose prediction.
Finally, we visualize and investigate the learned attention to obtain an intuition into how \modelname~is being used by the network.

\subsection{Camera Pose Prediction}
This experiment follows that of \cite{Wang20}.
The aim is to estimate the relative camera pose between pairs of images extracted at random from the MegaDepth test set. 
The pairs of images are divided into three subsets depending on the relative angular change: {\bf easy} ($[0^\circ, 15^\circ]$), {\bf medium} ($[15^\circ, 30^\circ]$), and {\bf hard} ($[30^\circ, 60^\circ])$.
Each subset has at least $1000$ pairs.

In order to determine the relative camera pose, we follow the approach of \cite{Wang20}. 
The essential matrix is extracted by using the mutual nearest neighbour correspondences and known camera intrinsic parameters.
The essential matrix is decomposed into the rotation and translation matrices.
The estimated angular change in rotation and translation is then compared to the ground truth.
If the difference between the predicted and ground truth is less than a threshold of $10^\circ$, the prediction is considered correct.

We consider two variants of injecting \modelname into the CAPSNet architecture. First, (C \modelname) only injects one \modelname at a coarse resolution.
Second, (C+F \modelname) injects two \modelname{}s at a coarse and a fine resolution.
We report the percentage of correct images for rotational and translational errors separately in \tabref{tab:megadepthresults}. 
These results demonstrate that using a \modelname{} does indeed improve over the baseline model, especially on the harder angle pairs.
Injecting further \modelname{}s does not substantially increase performance but it consistently performs better than the original CAPSNet model.
This demonstrates the value of using our \modelname{} to condition descriptors on both images.

\subsection{Visualisation of \modelname's Attention}
Finally, we visualize \modelname's predicted attention in \figref{fig:attentioncoamresults} to obtain an intuition of how the additional image is used to improve the learned descriptors.
We note that there is no clear a priori knowledge of what the model {\em should} attend to.
The attention module could find regions of similar texture but varying style in order to be invariant to the style.  Or the module could attend to the right location in the other image.
However, the qualitative results imply that the model is making use of the \modelname to attend to relevant regions.

Additionally, we quantify how invariant the descriptors are with the \modelname and without.
We use the sets of images in the HPatches benchmark that vary in illumination.
One image is kept fixed (the target) and the other varied (the query). We then evaluate how much the query image's descriptors vary from those of the target by computing the L1 error. 
Our descriptors differ on average by $0.30 \pm 0.14$, whereas \cite{Wang20}'s descriptors differ more, by $0.41 \pm 0.17$.
This validates that the \modelname increases the invariance of corresponding descriptors under large amounts of {\em scene shift}.

%% file: aachen_results.tex
\begin{table}
    \scriptsize
    \centering
    
  \captionof{table}{{\bf Aachen Day-Night \cite{Sattler18}.} Higher is better. Ours does comparably or better than other sota setups. $^*$~indicates the method was trained on the Aachen dataset.}
  \label{tab:aachenexperiments}

  \begin{tabular}{@{}lccc@{}}
            \toprule
             Method & \multicolumn{3}{c}{Threshold Accuracy} \\
             \cmidrule{2-4}
             & 0.25m ($2^{\circ}$) & 0.5m ($5^\circ$) & 5m ($10^\circ$) \\
             \cmidrule{1-4}
             
             Upright RootSIFT \cite{Lowe04} & 36.7 & 54.1 & 72.5 \\
             DenseSFM \cite{Sattler18}  & 39.8 & 60.2 & 84.7 \\
             Han+, HN++ \cite{Mishkin18,Mishchuk17}  & 39.8 & 61.2 & 77.6 \\
             Superpoint \cite{Detone18}  & 42.8 & 57.1 & 75.5 \\
             DELF \cite{Noh17}  & 39.8 & 61.2 & 85.7 \\
             D2-Net \cite{Dusmanu19}  & {\bf 44.9} & 66.3 & {\bf 88.8} \\
             R2D2* \cite{Revaud19} & {\bf 45.9} & 66.3 & {\bf 88.8} \\
             \midrule
             Ours w/o cond & 42.9 & 62.2 & 87.8  \\
             Ours w/o conf & 43.9 & 64.3 & 86.7  \\
             Ours & {\bf 44.9} & {\bf 70.4} & {\bf 88.8}  \\
             \cmidrule{1-4}
             Ours (E-B1)  & {\bf 44.9} & 68.4 & {\bf 88.8}  \\
             \bottomrule
  \end{tabular}

\end{table}

%% file: sfmresults_stylization.tex
\begin{table}[t]
    \centering
    \caption{{\bf SfM.} We compare our approach to using SIFT features on 3D reconstruction. $\uparrow$: higher is better. $\downarrow$: lower is better.}
    \scriptsize
    
    \begin{tabular}{@{}crccccccccccccc@{}}
    \toprule
                & & \multicolumn{3}{c}{Large SfM}  \\
                \cmidrule{3-5} 
         \multicolumn{2}{r}{\bf Landmark:} &  Madrid Met.  & Gen. & Tow. of Lon.  \\

    \midrule
    \multirow{2}{*}[0pt]{\# Reg.~Ims $\uparrow$} & {\bf SIFT \cite{Lowe04}:} & 500 & 1035 & 804 \\
        & {\bf Ours:} & {\bf 702} & {\bf 1072} & {\bf 967} \\ \midrule
    \multirow{2}{*}[0pt]{\# Sparse Pts $\uparrow$} & {\bf SIFT \cite{Lowe04}:} & 116K & 338K & 239K \\
    & {\bf Ours: } & {\bf 256K} & {\bf 570K} & {\bf 452K}  \\ \midrule
    
    \multirow{2}{*}[0pt]{Track Len $\uparrow$} & {\bf SIFT \cite{Lowe04}:} & {\bf 6.32} & 5.52 & {\bf 7.76}  \\
    & {\bf Ours: } & 6.09 & {\bf 6.60} & 5.82  \\ \midrule
    
    \multirow{2}{*}[0pt]{Reproj Err (px) $\downarrow$} & {\bf SIFT \cite{Lowe04}:} & {\bf 0.60} & {\bf 0.69} & {\bf 0.61}  \\
    & {\bf Ours: } & 1.30 & 1.34 & 1.32  \\ \midrule
    
    \multirow{2}{*}[0pt]{\# Dense Pts $\uparrow$} & {\bf SIFT \cite{Lowe04}:} & {\bf 1.8M} & {\bf 4.2M} & {\bf 3.1M}  \\
    & {\bf Ours: } & 1.1M & 2.1M & 1.8M  \\

    \bottomrule
    \end{tabular}
    \label{tab:reconstructionresults}
\end{table}

%% file: stylisation_results_stylisation.tex
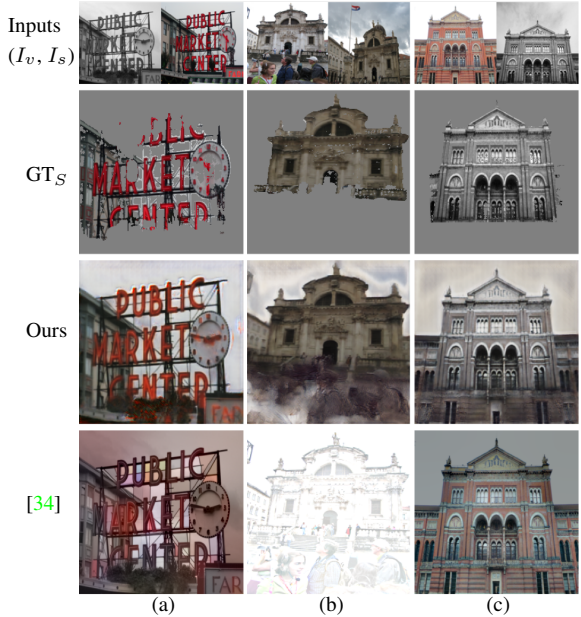
\begin{figure}
    \begin{minipage}{\linewidth}
        \centering
        \begin{overpic}[width=0.8\linewidth]{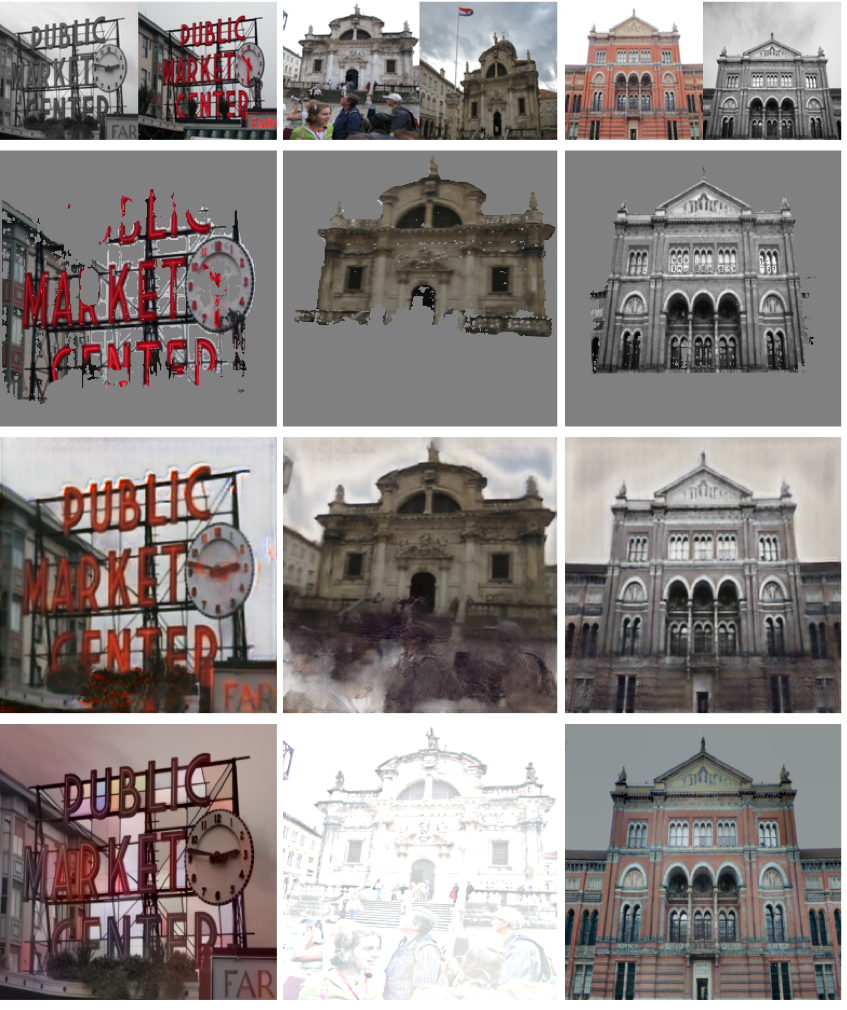}
        \put(-9,15){\footnotesize \cite{Luan17}}
        \put(-9,44){\footnotesize Ours}
        \put(-9,70){\footnotesize GT$_S$}
        \put(-12,95){\footnotesize Inputs}
        \put(-12,90){\footnotesize ($I_v$, $I_s$)}
        \put(12,-2){\footnotesize (a)}
        \put(40,-2){\footnotesize (b)}
        \put(68,-2){\footnotesize (c)}
        \end{overpic}
        \vspace{0.5em}
    \end{minipage} 
        \caption{{\bf Stylization.} Given $I_s$ and $I_v$, the task is to generate an image with the pose and viewpoint of $I_v$ and style of $I_s$. We show results for \modelname~and a baseline that uses semantics \cite{Luan17}. We also show the resampled image (GT$_S$) which is computed using the true correspondences from the MegaDepth dataset \cite{Li18}.
        While \cite{Luan17} works well for easy cases, it sometimes copies style from $I_v$ (as shown by the red background in (a) and red hued building in (c)). \cite{Luan17} also fails if the semantic prediction is incorrect (e.g.~(b)).}
        \label{fig:stylisationresults}
\end{figure}

%% file: attentioncoam_stylization.tex
\begin{figure}[t]
    \begin{minipage}{\linewidth}
        \centering
        \begin{overpic}[width=0.8\linewidth]{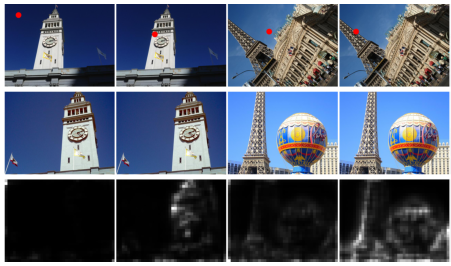}
        \put(-5,48){\footnotesize $I^1$}
        \put(-5,28){\footnotesize $I^2$}
        \put(-5,8){\footnotesize $A$}
        \end{overpic}
        \vspace{0.5em}
    \end{minipage} 
        \caption{{\bf CoAM Attention.} We visualize the predicted attention ($A$) for sample image pairs. The red dot in $I^1$ denotes the point for which we compute the attention. It is not clear apriori what the attention module should do, but it does attend to relevant, similar regions in the other image and is dependent on the query location.}
        \label{fig:attentioncoamresults}
\end{figure}

%% file: megadepth_results_ss.tex
\begin{table}[t]
    \centering
    \footnotesize
    \begin{tabular}{c c c c} \toprule
         {\bf Method} & \multicolumn{3}{c}{\bf Accuracy on MegaDepth} \\
         & {\em easy} & {\em medium} & {\em hard} \\ \toprule
         CAPS \cite{Wang20} w/ SIFT Kp. & 91.3 / {\bf 52.7} & 82.5 / 57.9 & 65.8 / 61.3  \\ \midrule
         Ours (C \modelname{})                           & 91.7 / 52.1 & {\bf 82.9} / {\bf 58.6} & {\bf 69.3} / 62.4   \\
         Ours  (C+F \modelname{}s)         & {\bf 91.9} / 52.3 & 82.8 / 58.4 & 68.8 / {\bf 63.4}  \\ \bottomrule
    \end{tabular}
    \caption{Results on the MegaDepth dataset on three increasingly challenging subsets ({\em easy}, {\em medium}, and {\em hard}) for both angular / translational errors: ($\cdot$) / ($\cdot$).
    The results show that augmenting the baseline model with our \modelname{}s improves performance, especially on the challenging viewpoint images, demonstrating the utility of conditioning the descriptors on {\em both} images under these conditions. }
    \label{tab:megadepthresults}
\end{table}

%% file: conclusion.tex
\section{Conclusion}

We investigated a new approach for obtaining correspondences for image pairs using a co-attention module and distinctiveness score.
The central insight was that, using neural networks, descriptors can be conditioned on {\em both} images.
This allows greater flexibility, as the descriptors only need to be invariant to changes between the pair of images.
Using this insight, our simple model improved the quality of the learned descriptors over those of a baseline model on multiple tasks and in both a supervised and self-supervised setting. We would expect further improvements with larger, more complex models.

\noindent {\bf Acknowledgements.}
The authors thank Noah Snavely and David Murray for
feedback and discussions. This work was funded by an EPSRC studentship, EPSRC
Programme Grant Seebibyte EP/M013774/1, a Royal Society Research Professorship, and ERC 638009-IDIU.

%% file: appendix.tex
\section{Overview}

We include additional implementation details and results in \secref{sec:othersetups}.
These experiments demonstrate the following.
First, the importance of both the distinctiveness score and \modelname to achieve sota results with comparison to other single stage approaches on the challenging YFCC100 dataset.
Second, they demonstrate how our approach can use a refinement step to achieve state-of-the-art results for all thresholds $>$1px on HPatches.
Third, they provide additional ablations including the use of a Noise Contrastive (NCE) loss as opposed to the hinge loss presented in the paper.
Finally, we provide more results and explanation for the SfM results given in the main paper.
We define the metrics used in obtaining the SfM results in \secref{sec:sfmterminology} and provide additional baselines.
We also demonstrate that our model is more robust than the one using SIFT on a challenging, new dataset of Henry Moore sculptures.
Finally, we provide visualisations of our reconstructed 3D models here and in the attached video.

Additionally, we provide further details of the architectures used in \secref{sec:architecture} for the \networkname, \modelname + CAPSNet \cite{Wang20}, and the stylization model.

We also provide qualitative samples of the distinctiveness scores learned by our model in \secref{sec:architecturechoices}.
We provide additional qualitative results for the stylization experiment, HPatches dataset, SfM experiment, and Aachen dataset in \secref{sec:quantitativeexperiments}.

\section{Additional Experiments}
\label{sec:othersetups}
In this section we report the results of additional experiments which consider additional ablations of the grid size chosen at test time (\secref{sec:ablategrid}), a refinement of our model to improve local matching (\secref{sec:reflocalmatching}) to achieve state-of-the-art results on HPatches, further comparisons of our model on the 3D reconstruction task (\secref{sec:sfmfurther}),   results on the YFCC100M dataset (\secref{sec:yfcc}), and results using another popular contrastive loss (NCE) (\secref{sec:nce}).

\subsection{Further Ablations}
\label{sec:ablategrid}
In this section we discuss and ablate how we select candidate matches at test time.

In order to compare all descriptor vectors at test time, we operate as follows. 
We create a $G \times G$ pixel grid and bilinearly interpolate on this grid from both the descriptor maps and distinctiveness scores. (Note that we have to normalize the interpolated descriptors.)
We consider all pairs of descriptors as candidate matches and
compare all descriptors in one image to those in the other.
In practice we use $G=128$.

At test time, we could use a larger grid size for better granularity.
However, this comes with the additional computational cost of performing $G^4$ comparisons. 
We tried using a larger grid ($G=256$) on the Aachen-Day Night dataset in \tabref{tab:aachenexperiments_nce_supp} but obtained comparable results to using $G=128$.
As a result, we continued to use $G=128$ for all our experiments.
Using a larger grid for the larger datasets in SfM (where the number of images are approximately $1500$) would have made these experiments intractable using current pipelines. 

However, we note that because we only consider points on a grid, we are losing some granularity which presumably impacts the performance on the 3D reconstruction tasks.
We discuss a method to refine the correspondences to obtain a finer granularity in the next section and the resulting complications.

\input{hpatches_results_refine}

\subsection{Refining Local Matching}
\label{sec:reflocalmatching}
In this section, we demonstrate that our local matching results can be improved by using a simple refinement strategy.

{\bf Description of Refinement Strategy.}
In order to refine the matches obtained using \networkname, we use a local neighbourhood to refine the match in the second image.
Given a correspondence with location $(x_1,y_1)$ in the first image and $(x_2,y_2)$ in the second image, we look at the similarity score between $(x_1,y_1)$ and the locations in a 3x3 grid centered on  $(x_2,y_2)$.

These scores are used to reweight the location in the second image using a normalized weighted sum with one difference.
Because the similarity scores are not evenly distribute and cluster around 0.5, they give too much weight to less likely matches. As a result, we subtract the minimum similarity from all scores in a local neighbourhood before computing the normalized weighted sum.

{\bf Results.}
The results are given in \figref{fig:hpatches_refine_supp}.
As can be seen, this simple refinement scheme gives a large boost in performance.
In particular, using this simple modification gives superior results to state of the art approaches for pixel thresholds $>4$px for viewpoint and all thresholds $>2$px for illumination. Overall, our model with the refinement scheme achieves state-of-the-art results for all thresholds $>1$px.

{\bf Discussion.}
While we could achieve high quality performance on HPatches using this simple modification, we note that it is not straightforward to apply this to the camera localization or 3D reconstruction tasks.
This is because both of these tasks require a single point to be tracked over multiple images in order to perform 3D reconstruction (the camera localization performs 3D reconstruction as part of the evaluation pipeline).
3D reconstruction pipelines assume a detect and describe pipeline for extracting matches, which implicitly have this assumption baked in to their setup, as they match the same detected points across different images.

However, this assumption is not implicit to our approach, as we only find correspondences between pairs of images at a time.
Further refining the points means that the location of a refined point from one pair of images will not necessarily match that of another pair, invalidating the track and negatively affecting the 3D reconstruction.
Incorporating these refined points would require rethinking how we incorporate refined correspondences in a 3D reconstruction pipeline. 
For example, we could use a reference image against which further images are compared and incorporated. 
Once a reference image has been exhausted, a new reference image would be chosen and so on.
We leave such an investigation to future work, but the boost in performance demonstrates the further potential of our setup.

\subsection{SfM Terminology}
\label{sec:sfmterminology}

Here we define the metrics used in reporting the SfM results. Note that these metrics are only {\em indicative} of the quality of the 3D model; please look at the reconstructed models in the zipped video for a qualitative estimate as to their respective quality.

\begin{enumerate}
    \item $\uparrow$ \# Reg. Ims: {\bf The number of registered images.} This is the number of images that are able to be put into correspondence and for which cameras were obtained when doing the 3D reconstruction. A higher values means more images were registered, implying a better model.
    \item $\uparrow$ \# Sparse Pts: {\bf The number of sparse points.} This is the number of sparse points obtained after performing the 3D geometry estimation. The higher the number indicates a better model, as more correspondences were able to be triangulated.
    \item $\uparrow$ Track Len: {\bf The track length.} How many images a given 3D point is seen in on average. If this is higher, it indicates that the model is more robust, as more images see that 3D point.
    \item $\downarrow$ Reproj err: {\bf The repreojection error.} This is the average pixel error between a 3D point and its projection in the images. If this is lower, it indicates the 3D points are more accurate.
    \item $\uparrow$ \# Dense Points: {\bf The number of dense points.} This is the number of dense points in the final 3D model. The higher this is, the more of the 3D structure was able to be reconstructed.
\end{enumerate}

\input{sfmresults_supp}

\subsection{Further Results for SfM}
We include additional baselines on the Local Feature Evaluation Benchmark of the original paper.
We additionally include results in a challenging scenario where the dataset contains fewer  ($\approx 10-100$ images) images of the 
same scene and where the object (a sculpture) may differ in material, location, and context.

\subsubsection{Further Baselines on Local Feature Evaluation Benchmark}
\label{sec:sfmfurther}
In this section we compare our 3D reconstruction on the Local Feature Evaluation Benchmark \cite{Schoenberger17} to two additional baselines \cite{Dusmanu19,Luo18} in \tabref{tab:sfmresults_supp}.
These results were not included in the paper as these additional baselines perform similarly to SIFT \cite{Lowe04} and there was limited space.
However, we note that both of these baselines use learned descriptors, yet they do not perform any better than SIFT in terms of the number of registered images and sparse 3D points.
Our method performs significantly better across all three large scale datasets for obtaining sparse 3D points and registering images.

\subsubsection{Further Results on a Sculpture SFM Benchmark}
\label{sec:henrymoore}
We use images from the Sculpture dataset \cite{Fouhey16}, which consists of images of the same sculpture downloaded from the web.
As an artist may create the same sculpture multiple times, a sculpture's  material (e.g.~bronze or marble), location, or context (e.g.~the season) may change in the images (refer to the supplementary for examples).
In particular, we evaluate on nine sculptures by the artist Henry Moore.
These sets of images contain large variations and the sculpture itself is often smooth, leading to less texture for finding repeatably detectable regions.

We report the results in \tabref{tab:henrymooreresults} and visualise samples in \figref{fig:sfm_supp}.
While these metrics are proxies for reconstruction accuracy, our approach is able to consistently obtain more 3D points than the others for each image set.
These results validate that our approach does indeed make our model robust in this context and it performs as well if not better than the SIFT baseline method.

\input{yfcc_results}

\subsection{The YFCC100M Dataset}
\label{sec:yfcc}
Here we report results for our model and ablations on the YFCC100M \cite{Thomee16} dataset.
This dataset further demonstrates the superiority of our approach to other detect and describe setups and the utility of each component of our model (i.e.~the distinctiveness score and \modelname).

{\bf Setup.}
The task of this dataset is to perform two view geometry estimation on four scenes with 1000 pairs each.
Given a pair of images, the task is to use estimated correspondences to predict the essential matrix using the known intrinsic matrices.
The essential matrix is decomposed into the rotation and translation component \cite{Hartley00}.
The reported error metric is the percentage of images that have the rotation and translation error (in degrees) less than a given threshold.

To run \networkname~on this dataset, we first use \networkname~to extract high quality matches for each pair of images.
We use the known intrinsics to convert these to points in camera space.
We then use RANSAC \cite{Fischler81} and the 5-point algorithm in order to obtain the essential matrix \cite{Hartley00}.

{\bf Results.}
The results are given in \tabref{tab:yfcc_results}.
They demonstrate that our model achieves superior performance for low error thresholds to other methods that directly operate on extracted features.
The results further demonstrate that the distinctiveness score and \modelname are crucial for good performance, validating our model design.

Finally, our model does a bit worse than RANSAC-Flow \cite{Shen20}.
However, we note \cite{Shen20} uses a segmentation model to restrict correspondences to only regions in the foreground of the image (e.g.~the segmentation model is used to remove correspondences in the sky). 
Additionally, this method first registers images under a homography using predetected correspondences and then trains a network on top of the transformed images to perform fine-grained optical flow. 
As a result, this method performs as well as the underlying correspondences. 
Considering that our method has consistently been demonstrated to perform comparably or better than previous approaches for obtaining correspondences, we note that this method could be used on top of ours for presumably further improved performance.
However, as this work was only recently published, we leave this for future work.

\input{figs_supp_quant.tex}

\subsection{An InfoNCE Loss}
\label{sec:nce}
In the paper we demonstrate the robustness of our approach to the precise choice of architecture (e.g.~we can achieve impressive results using a ResNet \cite{He15} or EfficientNet \cite{Tan19} backbone). 

Here, we consider using the CPC objective \cite{Oord18}, which is inspired by Noise-Contrastive Estimation (NCE) approaches \cite{Gutmann10,Mnih13b}, and so is called an InfoNCE loss. While this is also a contrastive loss, similarly to the hinge loss in the main paper, the implementation is different. We find that we can still achieve impressive results with this loss, demonstrating the robustness of the approach to the precise choice of loss. 

\subsubsection{Implementation}
We follow the implementation of \cite{Oord18}, except that because we use normalized features, we add a temperature $\tau$. This is essential for good performance.

The setup is the same as that described in the paper, except for the implementation of the loss.
Assume we have two descriptor maps $D^1$ and $D^1$ corresponding to the two input images $I^1$ and $I^2$.
At a location $i$ in $D^1$, we obtain the descriptor vector $d^1_{i} \in \mathbbm{R}^c$.
To compare descriptor vectors, we first normalize and then use cosine similarity to obtain a scalar matching score:
\begin{equation}
\label{eq:corrloss}
    s(d^1_i, d^2_j) = \left( \frac{d^1_{i}}{||d^1_{i}||_2} \right)^T \frac{d^2_j}{||d^2_j||_2}.
\end{equation}
If the score is near $1$, this is most likely a match. If it is near $-1$, it is most likely not a match.

Again, as in the paper, given two images of a scene $I^1$ and $I^2$ with a known set of correspondences from MegaDepth \cite{Li18}, we randomly select a set $\mathcal{P}$ of  $L$ true correspondences.
For each positive correspondence $p$, we additionally select a set $\mathcal{N}_p$ of  $N$ negative correspondences.
The loss $\mathcal{L}_{\texttt{nce}}$ is then
\begin{equation}
     - \log \frac{1}{L} \sum_{p=(x,y) \in \mathcal{P}} \frac{e^{(\tau * s(d^1_{x}, d^2_{y})}}{e^{(\tau * s(d^1_{x}, d^2_{y}))} + \sum_{(x,\hat{y}) \in \mathcal{N}_p} e^{(\tau * s(d^1_{x}, d^2_{\hat{y}}))}  }
\end{equation}
where $\tau=20$ is a temperature.

\subsubsection{Experiments}
We train the InfoNCE model using the $\mathcal{L}_{\texttt{nce}}$ in the same manner as in the paper and evaluate it on two of the datasets discussed in the paper: HPatches \cite{Balntas17} and Aachen \cite{Sattler18, Sattler12}.

{\bf HPatches.}
The results are given in \figref{fig:hpatches_nce_supp}.
From here we see that our model with an NCE loss performs competitively on this dataset, obtaining superior results to that of the model in the paper. 

{\bf Aachen Day-Night.}
The results are given in \tabref{tab:aachenexperiments_nce_supp}.
These results demonstrate that using an NCE loss with our backbone achieves results competitive with other state-of-the-art approaches but it performs a bit worse than the hinge loss used in the paper.

{\bf Discussion.}
These experiments have shown that we can achieve high quality results when using a different but effective contrastive loss. 
As a result, our approach is robust to not only the backbone architecture (as shown in the paper) but also the precise choice of the contrastive loss.

\section{Architectures}
\label{sec:architecture}

\input{architecture}

\subsection{Architecture for \networkname}
The components of the main model are described in the main text.
Here we give further details of the different components.
The encoder is a ResNet50 model, except that we extract the features from the last two blocks to obtain feature maps $f^i_S$ and $f^i_L$.
The details are given in \tabref{tab:encoder}.

The features $f^i_L$ and $f^i_S$ are projected in order to reduce the number of channels using linear layers.
There are four linear layers (one for each of $f^1_L$, $f^2_L$, $f^1_S$, $f^2_S$).
The linear layers operating at the larger resolution ($f^i_L$) project the features from $2048$ size vectors to $256$.
The linear layers operating at the smaller resolution ($f^i_S$) project the features from $1024$ size vectors to $128$.

The decoder consists of a sequence of decoder layers.
A layer takes the bi-linearly upsampled features from the previous layer, the corresponding encoded features, and optionally the attended features.
The details are given in \tabref{tab:decoder}.
Finally, the unnormalized features are passed to a MLP which regresses the distinctiveness score.
The MLP consists of three blocks of linear layer (with no bias) and batch normalization followed by a sigmoid layer. The channel dimensions are $64 \rightarrow 1 \rightarrow 1$.

\subsection{Architecture for \modelname + CAPSNet \cite{Wang20}}
Here we further describe the CAPSNet architecture and how we incorporate \modelname{}s into the architecture.
We use the ResNet34 encoder (in order to fit a batch size of $6$ on our GPUs).
We extract the features from the 2nd and 3rd blocks to obtain feature maps $f^i_L$ (at a fine level) and $f^i_S$ (at a coarse level).
The details are given in \tabref{tab:encodercapsnet}.

The features $f^1_L$, $f^2_L$, $f^1_S$, and $f^2_S$ are projected as above.
The linear layers operating at the larger resolution ($f^i_L$) project the features from $128$ size vectors to $16$.
The linear layers operating at the smaller resolution ($f^i_S$) project the features from $256$ size vectors to $32$.

The decoder operates as above.
The details are given in \tabref{tab:decodercapsnet}.
This gives the final set of descriptors of size 128D.
We find that using just the fine features performs (output of deconv\_1) better than usin a concatenation of the coarse (at a resolution $60 \times 80$ and fine features).

\subsection{Architecture for Stylization}
In \figref{fig:architecture_supp}, we illustrate further our method for stylizing images using our initial set of dense correspondences. Given two images $I_v$ and $I_S$, the task is to generate an image with the viewpoint of $I_v$ and the style of $I_S$. In brief, we first use the dense correspondences to sample from $I_S$ to obtain the initial image. We then use a refinement network to fix errors and fill in missing regions. We do this in two stages. The first stage fixes errors and can be trained with an L1 loss. However, we wish to use a discriminator loss, but using this directly on the intermediary image causes information to leak and the generated image to match $I_v$, which is input to the network.
To use a discriminator loss without leaking information, we use a second network (a sequence of ResNet blocks) which is trained with both an L1 and discriminator loss. Crucially, we do not allow gradients to flow from the second network (the set of ResNet blocks) to the first.

\input{archs_supp}

\section{Additional Results}

\subsection{Further Visualisation of \modelname's Attention}
In \figref{fig:attention_supp} we illustrate additional predicted attention maps from our \modelname when trained with CAPSNet \cite{Wang20} in a self-supervised framework.
Again, it is not clear apriori what the model {\em should} attend to. 
However, in these results, we can see how the attention varies as a function of the query location and seems to attend to relevant regions.

\subsection{Visualising the Distinctiveness Score}
\label{sec:architecturechoices}
In \figref{fig:conf_supp} we illustrate the predicted distinctiveness score and in \figref{fig:similarityscore} the similarity score. Here we can see that the most confident parts of both images are regions that exist in {\em both} images. We further test this by looking at what the distinctiveness scores look like when we keep one input view the same but change the other in \figref{fig:conf_cond_supp}.
We can see that the distinctiveness score changes depending on the input images: the output is indeed dependent on {\em both} input images.

To visualise the similarity score, we proceed as follows in \figref{fig:similarityscore}.
For a query point $k$ in $I_1$, we display the correspondence map $c_{kl}$. The point in the sky matches a region around the building; a point on the arch matches a point on the arch (shown by the bright spot)

\subsection{Qualitative Results}
\label{sec:quantitativeexperiments}

In \figref{fig:aachen_supp} we show random matches obtained by our method on random samples from the Aachen Day-Night test set and similarly in \figref{fig:hpatches_supp} for HPatches, \figref{fig:sfm_supp} for 3D reconstruction using SfM and \figref{fig:stylise_supp} for the stylization task.

  \input{figs_supp}

\clearpage

%% file: hpatches_results_refine.tex
\begin{figure}[t]
    \centering
    \includegraphics[width=\linewidth]{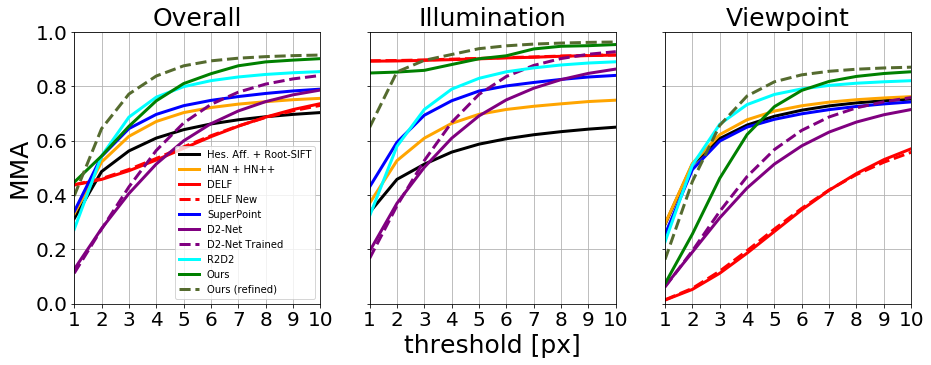}
    \caption{{\bf Results on HPatches when refining correspondences.} In this experiment, we look at the results of using a refinement strategy based on a local neighbourhood to improve the accuracy of the detected correspondences. We see that using the refinement scheme, we obtain a big boost in performance. In particular, we improve our results for fine-grained pixel thresholds on the viewpoint task. We achieve comparable results with state-of-the-art methods for small pixel thresholds and better performance for larger thresholds ($>4$px). On illumination we maintain performance except for very small thresholds ($\leq 1$px). This small degradation is probably due to limited noise that is introduced with the refinement strategy. In general, using this strategy we improve our results to achieve state of the art performance for all pixel thresholds.}
    \label{fig:hpatches_refine_supp}
\end{figure}

%% file: sfmresults_supp.tex
\begin{table*}[t]
    \centering
    \caption{{\bf SfM.} We compare our approach to three baselines on 3D reconstruction for three scenes with a large number of images. Our method obtains superior performance across the metrics except for reprojection error, despite using coarse correspondences and a single scale. In particular, our method registers more images and obtains more sparse 3D points. $\uparrow$ denotes higher is better. $\downarrow$ denotes lower is better. }
    \scriptsize
    \begin{tabular}{@{}llccccc@{}}
    \toprule
                {\bf LMark} & {\bf Method} & {\bf $\uparrow$ \# Reg.~Imgs} & {\bf $\uparrow$ \# Sparse Pts} & {\bf $\uparrow$ Track Len} & {\bf $\downarrow$ Reproj.~Err} & {\bf $\uparrow$ \# Dense Pts} \\
                \midrule
                {\bf Madrid} & RootSIFT \cite{Lowe04} & 500 & 116K & 6.32 & {\bf 0.60px} & {\bf 1.82M} \\
                {\bf Metropolis} & GeoDesc \cite{Luo18} & 495 & 144K & 5.97 & 0.65px & 1.56M \\
                1344 images & D2 MS \cite{Dusmanu19} & 495 & 144K & {\bf 6.39} & 1.35px & 1.46M \\ 
                            & Ours &                   {\bf 702} & {\bf 256K} & 6.09 & 1.30px & 1.10M \\ \midrule
                
                {\bf Gendarmen-} & RootSIFT \cite{Lowe04} & 1035 & 338K & 5.52 & {\bf 0.69px} & {\bf 4.23M} \\
                {\bf markt} & GeoDesc \cite{Luo18} & 1004 & 441K & 5.14 & 0.73px & 3.88M \\
                1463 images & D2 MS \cite{Dusmanu19} & 965 & 310K & 5.55 & 1.28px & 3.15M \\ 
                            & Ours &                   {\bf 1072} & {\bf 570K} & {\bf 6.60} & 1.34px & 2.11M \\ \midrule
                
                {\bf Tower of} & RootSIFT \cite{Lowe04} & 804 & 239K & {\bf 7.76} & {\bf 0.61px} & {\bf 3.05M} \\
                {\bf London} & GeoDesc \cite{Luo18} & 776 & 341K & 6.71 & 0.63px & 2.73M \\
                1576 images & D2 MS \cite{Dusmanu19} & 708 & 287K & 5.20 & 1.34px & 2.86M \\ 
                            & Ours &    {\bf 967}   & {\bf 452K} & 5.82 & 1.32px & 1.81M \\ 
                
    \bottomrule
    \end{tabular}
    \label{tab:sfmresults_supp}
\end{table*}

\begin{table*}[t]
    \centering
    \caption{{\bf SfM.} We compare our approach to using SIFT features on 3D reconstruction on a challenging new SFM dataset of sculptures. This dataset contains images from the web containing large variations in illumination and viewpoint. These metrics are a proxy for 3D reconstruction quality, so we encourage the reader to view the reconstructions in the supplementary. X: failure. $\uparrow$: higher is better. $\downarrow$: lower is better. }
    \scriptsize
    \begin{tabular}{@{}crcccccccccccc@{}}
    \toprule
                &  \multicolumn{9}{c}{Sculpture Dataset } \\
                 \midrule
         \multicolumn{2}{r}{\bf Landmark:} & HM1 & HM2 & HM3 & HM4 & HM5 & HM6 & HM7 & HM8 & HM9 \\
         \multicolumn{2}{r}{\# Images:} & 12 & 124 & 250 & 266 & 78 & 31 & 358 & 238 & 74 \\

    \midrule
    \multirow{2}{*}[0pt]{\# Reg.~Ims $\uparrow$} & {\bf SIFT \cite{Lowe04}:}  & X & 103 & {\bf 198} & 212 & 61 & 22 & 266 & {\bf 201} & 53  \\
        & {\bf Ours:} & {\bf 12} & {\bf 108} & 194 & {\bf 215} & {\bf 67} & {\bf 25} & {\bf 284} & {\bf 201} & {\bf 57}  \\ \midrule
    \multirow{2}{*}[0pt]{\# Sparse Pts $\uparrow$} & {\bf SIFT \cite{Lowe04}:} & X & 48K & 70K & 102K & 28K & 9K & 128K & {\bf 99K} & {\bf 23K} \\
    & {\bf Ours: } & {\bf 2.9K} & {\bf 63K} & {\bf 83K} & {\bf 121K} & {\bf 40K} & {\bf 10K} & {\bf 190K} & {\bf 99K} & 21K \\ \midrule
    
    \multirow{2}{*}[0pt]{Track Len $\uparrow$} & {\bf SIFT \cite{Lowe04}:} & X & {\bf 5.33} & {\bf 5.92} & {\bf 5.80} & {\bf 4.54} & {\bf 4.73} & {\bf 4.46} & {\bf 5.24} & 4.75 \\
    & {\bf Ours: }  & {\bf 3.60} & 5.03 & 5.43 & 5.61 & 4.32 & 4.00 & 4.23 & {\bf 5.24} & {\bf 4.77} \\ \midrule
    
    \multirow{2}{*}[0pt]{Reproj Err (px) $\downarrow$} & {\bf SIFT \cite{Lowe04}:} & X & 1.31 & 1.32 & {\bf 1.28} & 1.30 & 1.33 & {\bf 1.22} & {\bf 1.30} & {\bf 1.32} \\
    & {\bf Ours: } & {\bf 1.33} & {\bf 1.30} & {\bf 1.30} & 1.29 & {\bf 1.29} & {\bf 1.26} & 1.23 & {\bf 1.30} & {\bf 1.32} \\ \midrule
    
    \multirow{2}{*}[0pt]{\# Dense Pts $\uparrow$} & {\bf SIFT \cite{Lowe04}:}  & X & 160K & 143K & {\bf 307K} & 73K & {\bf 46K} & 174K & 333K & {\bf 54K} \\
    & {\bf Ours: } & {\bf 0.2K} & {\bf 188K} & {\bf 156K} & 296K & {\bf 82K} & 44K & {\bf 187K} & {\bf 333K} & 53K \\

    \bottomrule
    \end{tabular}
    \label{tab:henrymooreresults}
\end{table*}

%% file: yfcc_results.tex
\begin{table}[t]
    \centering
    \footnotesize

    \caption{Results on the YFCC dataset \cite{Thomee16}. Higher is better. Our approach outperforms all other detect and describe approaches (e.g.~all but RANSAC-Flow) that operate directly on features for smaller angle errors and performs competitively for larger angle errors. Additionally, this dataset clearly demonstrates the utility of \modelname~and the distinctiveness score. $^\dagger$Note that RANSAC-Flow \cite{Shen20} is a multi stage approach that iteratively registers images. Such an approach could be added on top of ours.}
    \label{tab:yfcc_results}
    
    \begin{tabular}{lcc} \toprule
        Method & mAP@$5^\circ$ & mAP@$10^\circ$  \\ \midrule
        SIFT \cite{Lowe04} & 46.83 & 68.03  \\
        Contextdesc \cite{Luo19} & 47.68 & 69.55 \\
        Superpoint \cite{Detone18} & 30.50 & 50.83 \\
        PointCN \cite{Moo18,Zhang19Feat} & 47.98 & - \\
        PointNet++ \cite{Qi17,Zhang18Feature} & 46.23 & - \\
        N$^3$Net \cite{Plotz18,Zhang18Feature} & 49.13 & -  \\
        DFE \cite{Ranft18,Zhang18Feature} & 49.45 & -  \\
        OANet \cite{Zhang18Feature} & 52.18 & -  \\
        RANSAC-Flow$^\dagger$ \cite{Shen20} & 64.68 & 73.31  \\
        \midrule
        Ours (w/o conf) & 31.60 & 40.80 \\
        Ours (w/o cond) & 53.43 & 65.13 \\ 
        Ours & 55.58 & 66.79 \\ 
        \cmidrule{1-3}
        Ours (E-B1) & {\bf 57.23} & {\bf 68.39}  \\
        \bottomrule
    \end{tabular}
\end{table}

%% file: figs_supp_quant.tex
\begin{figure}[t]
    \centering
    \includegraphics[width=\linewidth]{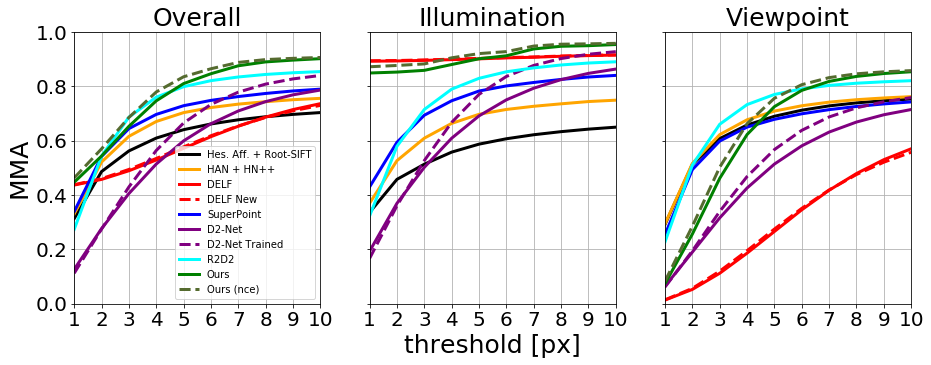}
    \caption{{\bf Results on HPatches using a NCE loss.} In this experiment, we look at the results of using a NCE loss as opposed to a hinge loss. We see that using a NCE loss, we still achieve high quality results, demonstrating the robustness of our approach. }
    \label{fig:hpatches_nce_supp}
\end{figure}

\begin{table}[]

    \caption{{\bf Results on Aachen Day-Night using a NCE loss.}  We can see that training with NCE is slightly worse than our hinge loss, but it is competitive with other state-of-the-art methods. Higher is better. * indicates the method was trained on the Aachen dataset. }
    \label{tab:aachenexperiments_nce_supp}
    
    \footnotesize
    
    \centering \begin{tabular}{@{}lcccc@{}}
            \toprule
             Method & Type & \multicolumn{3}{c}{Threshold Accuracy} \\
             \cmidrule{3-5}
             & & 0.25m ($2^{\circ}$) & 0.5m ($5^\circ$) & 5m ($10^\circ$) \\
             \cmidrule{1-5}
             
             Upright RootSIFT \cite{Lowe04} & Spa & 36.7 & 54.1 & 72.5 \\
             DenseSFM \cite{Sattler18} & Den & 39.8 & 60.2 & 84.7 \\
             Han+, HN++ \cite{Mishkin18,Mishchuk17} & Spa & 39.8 & 61.2 & 77.6 \\
             Superpoint \cite{Detone18} & Spa & 42.8 & 57.1 & 75.5 \\
             DELF \cite{Noh17} & Spa & 39.8 & 61.2 & 85.7 \\
             D2-Net \cite{Dusmanu19} & Spa & {\bf 44.9} & 66.3 & {\bf 88.8} \\
             R2D2* \cite{Revaud19} & Spa & {\bf 45.9} & 66.3 & {\bf 88.8} \\
             \midrule
             Ours (nce) & Den & 42.9  & 62.2 & 87.8   \\
             Ours ($G=256$) & Den & {\bf 44.9}  &  68.4 & 87.8   \\
             Ours & Den & {\bf 44.9} & 70.4 & {\bf 88.8}  \\
             \bottomrule
    \end{tabular}
\end{table}

%% file: architecture.tex
\begin{table}[t]
    \centering
    \scriptsize
    \caption{{\bf Encoder of \networkname.} The encoder is a Resnet50 \cite{He15} encoder. The convolutions column denotes the convolutional and max-pooling operations. Implicit are the BatchNorm and ReLU operations that follow each convolution.}
    \label{tab:encoder}
    
    \begin{tabular}{ccc}
    \toprule
         {\bf layer name} & {\bf output size}       & {\bf convolutions} \\ \midrule
         conv 1                         & $128 \times 128$                      & $7 \times 7, 64, $ stride $2$  \\ \midrule
         \multirow{2}{*}[0pt]{conv2\_x}  & \multirow{2}{*}[0pt]{$64 \times 64$}  & $3 \times 3$ max pool, stride $2$ \\ 
                                        &                                       & $\begin{pmatrix} 1 \times 1, 64  \\ 3 \times 3, 64 \\ 1 \times 1, 256 \end{pmatrix} \times 3$ \\ \midrule
        conv3\_x  & $32 \times 32$  & $\begin{pmatrix} 1 \times 1, 128  \\ 3 \times 3, 128 \\ 1 \times 1, 512 \end{pmatrix} \times 4$ \\ \midrule
        conv4\_x ($f^i_L$) & $16 \times 16$  & $\begin{pmatrix} 1 \times 1, 256  \\ 3 \times 3, 256 \\ 1 \times 1, 1024 \end{pmatrix} \times 6$ \\ \midrule
        conv5\_x ($f^i_S$) & $8 \times 8$  & $\begin{pmatrix} 1 \times 1, 512  \\ 3 \times 3, 512 \\ 1 \times 1, 2048 \end{pmatrix} \times 3$ \\ \bottomrule
    \end{tabular}
\end{table}

\begin{table}[t]
    \centering
    \scriptsize
    \caption{{\bf Decoder of \networkname.} The decoder is a UNet \cite{Ronneberger15} variant. The convolutions column denotes the convolutional operations. Implicit are the BatchNorm and ReLU operations that follow each convolution as well as the bi-linear upsampling operation that resizes features from the previous layer before the convolutional blocks.}
    \label{tab:decoder}
    
    \begin{tabular}{cccc}
    \toprule
         {\bf layer name} & {\bf inputs}            &{\bf output size} & {\bf convolutions} \\ \midrule
         deconv\_5         & conf5\_x ($\times 2$), $\hat{f}^i_S$ & $ 16 \times 16$  &  $\begin{pmatrix} 3 \times 3,256 \\
                                                                                           3 \times 3,256 \end{pmatrix}$ \\ \midrule
         deconv\_4         & deconv\_5, conv4\_x, $\hat{f}^i_L$ & $ 32 \times 32$  &  $\begin{pmatrix} 3 \times 3,256 \\
                                                                                           3 \times 3,256 \end{pmatrix}$ \\ \midrule
         deconv\_3         & deconv\_4, conv3\_x & $ 64 \times 64$  &  $\begin{pmatrix} 3 \times 3,128 \\
                                                                                           3 \times 3,128 \end{pmatrix}$ \\ \midrule
         deconv\_2         & deconv\_3, conv2\_x & $ 128 \times 128$  &  $\begin{pmatrix} 3 \times 3,128 \\
                                                                                           3 \times 3,128 \end{pmatrix}$ \\ \midrule
         deconv\_1         & deconv\_2, conv1\_x & $ 256 \times 256$  &  $\begin{pmatrix} 3 \times 3,64 \\
                                                                                           3 \times 3,64 \end{pmatrix}$ \\ \bottomrule
    \end{tabular}
\end{table}

\begin{table}[t]
    \centering
    \scriptsize
    \caption{{\bf Encoder of CAPSNet \cite{Wang20} variant.} The encoder is a ResNet34 \cite{He15} encoder. The convolutions column denotes the convolutional and max-pooling operations. Implicit are the BatchNorm and ReLU operations that follow each convolution.}
    \label{tab:encodercapsnet}
    
    \begin{tabular}{ccc}
    \toprule
         {\bf layer name} & {\bf output size}       & {\bf convolutions} \\ \midrule
         conv 1                         & $240 \times 320$                      & $7 \times 7, 64, $ stride $2$  \\ \midrule
         \multirow{2}{*}[0pt]{conv2\_x}  & \multirow{2}{*}[0pt]{$120 \times 160$}  & $3 \times 3$ max pool, stride $2$ \\ 
                                        &                                       & $\begin{pmatrix} 3 \times 3, 64  \\ 3 \times 3, 64 \end{pmatrix} \times 2$ \\ \midrule
        conv3\_x ($f^i_L$)  & $60 \times 80$  & $\begin{pmatrix} 3 \times 3, 128  \\ 3 \times 3, 128  \end{pmatrix} \times 2$ \\ \midrule
        conv4\_x ($f^i_S$) & $30 \times 40$  & $\begin{pmatrix} 3 \times 3, 256  \\ 3 \times 3, 256 \end{pmatrix} \times 2$ \\ \bottomrule
    \end{tabular}
\end{table}

\begin{table}[t]
    \centering
    \scriptsize
    \caption{{\bf Decoder of CAPSNet \cite{Wang20}.} The decoder is a UNet \cite{Ronneberger15} variant. The convolutions column denotes the convolutional operations. Implicit are the BatchNorm and ReLU operations that follow each convolution as well as the bi-linear upsampling operation that resizes features from the previous layer before the convolutional blocks.}
    \label{tab:decodercapsnet}
    
    \begin{tabular}{cccc}
    \toprule
         {\bf layer name} & {\bf inputs}            &{\bf output size} & {\bf convolutions} \\ \midrule
         deconv\_3         & $\hat{f}^2_S$, $f^1_S$ & $ 60 \times 80$  &  $\begin{pmatrix} 1 \times 1, 256 \\ 3 \times 3, 512 \end{pmatrix}$ \\ \midrule
         deconv\_2         &  $\hat{f}^2_L$, $f^1_L$, deconv\_3 & $ 120 \times 160$  &  $\begin{pmatrix} 1 \times 1, 256 \\ 3 \times 3, 512 \\ 3 \times 3, 256 \end{pmatrix}$ \\ \midrule
         deconv\_1         & deconv\_2, conv2\_x & $ 120 \times 160$  &  $\begin{pmatrix} 3 \times 3, 512 \\ 3 \times 3, 128 \end{pmatrix}$ \\ \midrule
    \end{tabular}
\end{table}

%% file: archs_supp.tex
\begin{figure*}
    \centering
    \includegraphics[width=\linewidth]{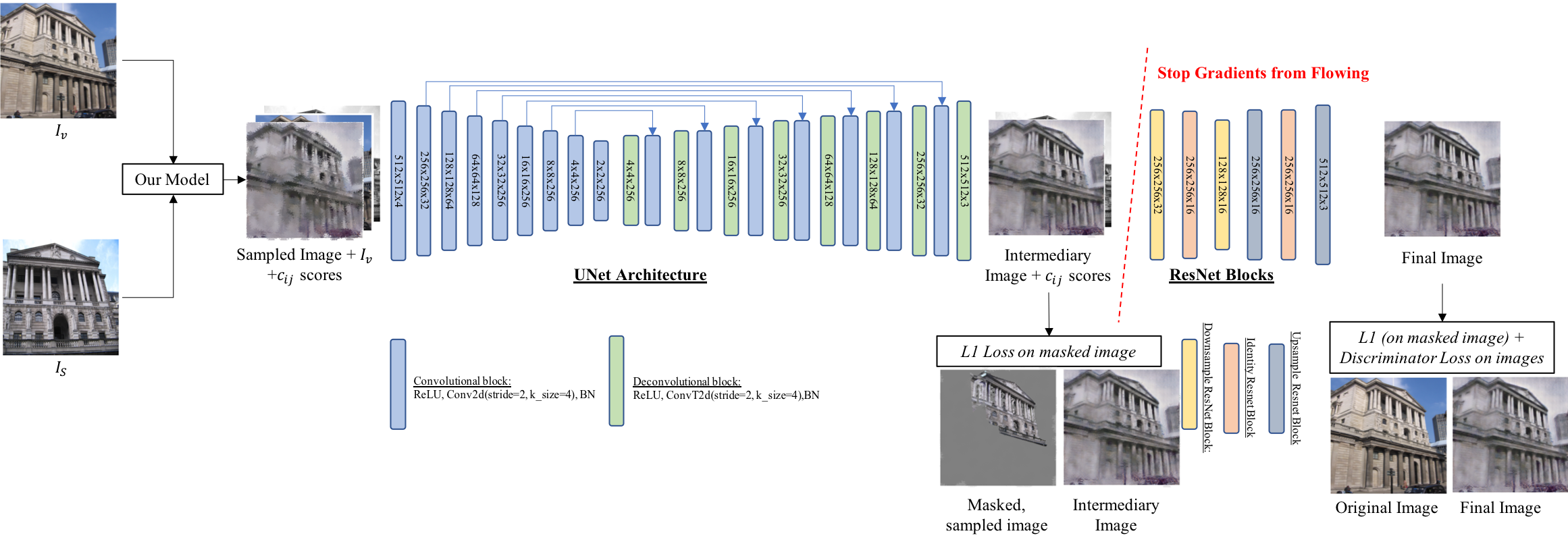}
    \caption{Illustration of the architecture used for the downstream stylization task. We first use our model to predict correspondences to transform the input image $I_S$ into the position of $I_v$. The next step is to learn how to fill in and fix errors with a discriminator. However, we additionally can train the portion of the generated image visible in both input images to match the true transformed image. We also can use high frequency information in $I_v$ when performing this transformation. However, if we train end-to-end then information will leak from $I_v$ to the generated image. As a result, we train in two stages. We first generate an intermediary image using a UNet \cite{Ronneberger15} which is trained using an L1 loss on the generated intermediary image for regions that are in common between the two images (and for which we can determine what the true pixel colour should be). We input this intermediary image to a set of ResNet blocks to refine the original prediction: this is trained with both a discriminator (pix2pixHD \cite{Wang18}) and L1 loss. }
    \label{fig:architecture_supp}
\end{figure*}

%% file: figs_supp.tex
\begin{figure*}
	\centering
    \begin{minipage}{\linewidth}
        \centering
        \begin{overpic}[width=0.95\linewidth]{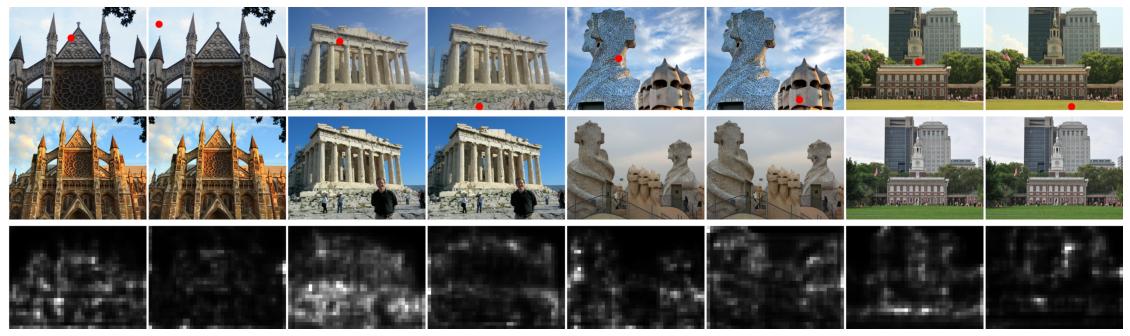}
        \put(-2,25){\footnotesize $I^1$}
        \put(-2,15){\footnotesize $I^2$}
        \put(-2,5){\footnotesize $A$}
        \end{overpic}
        \vspace{0.5em}
    \end{minipage} 
	\caption{{\bf \modelname Attention.} Here we visualize more samples of \modelname{}'s  predicted attention for sample image pairs. As in the main paper, the red dot in $I^1$ denotes the point for which we compute the attention. It is not clear apriori what the attention module should do but it does attend to relevant, similar regions in the other image and is dependent on the query location.}
	\label{fig:attention_supp}
\end{figure*}

\begin{figure*}
    \centering
    \includegraphics[width=\linewidth]{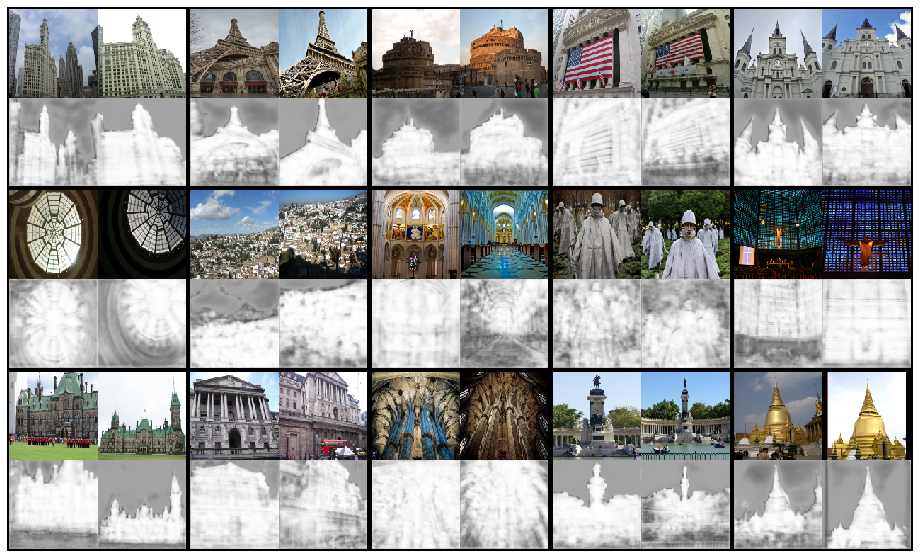}
    \caption{Random pairs associated with their predicted distinctiveness score on the MegaDepth test set. Top row shows the input pairs, bottom row the associated distinctiveness scores.}
    \label{fig:conf_supp}
\end{figure*}

\begin{figure*}
    \centering
    \includegraphics[width=0.2\linewidth]{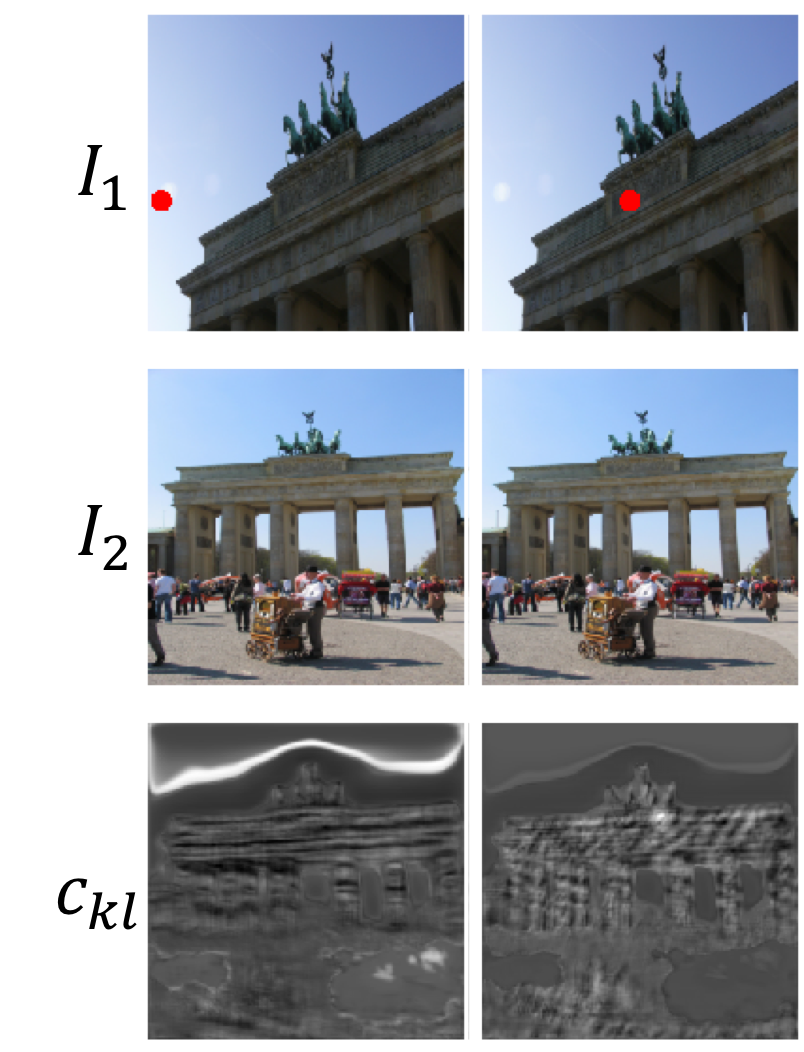}
    \caption{Pairs of images with their similarity scores on the MegaDepth test set. For a query point $k$ in $I_1$, we display the correspondence map $c_{kl}$. The point in the sky matches a region around the building; a point on the arch matches a point on the arch (shown by the bright spot).}
    \label{fig:similarityscore}
\end{figure*}

\begin{figure*}
    \centering
    \includegraphics[width=\linewidth]{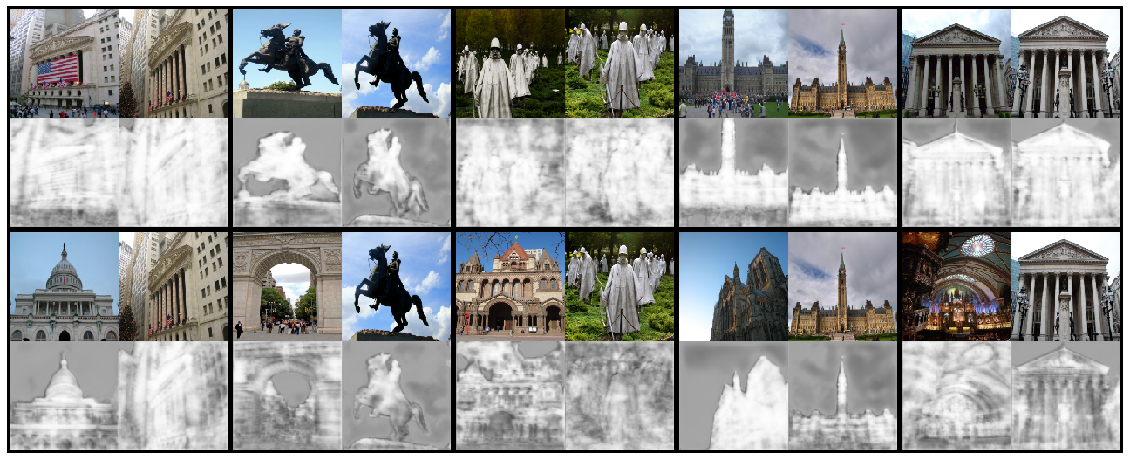}
    \caption{Random pairs associated with their predicted distinctiveness score on the MegaDepth test set. Top: Pairs of image of the same scene. Bottom: Pairs with one image from the top associated with another from a different scene. Notice that the distinctiveness score changes between the two cases and becomes irrelevant.}
    \label{fig:conf_cond_supp}
\end{figure*}

\begin{figure*}
    \centering
    \includegraphics[width=\linewidth]{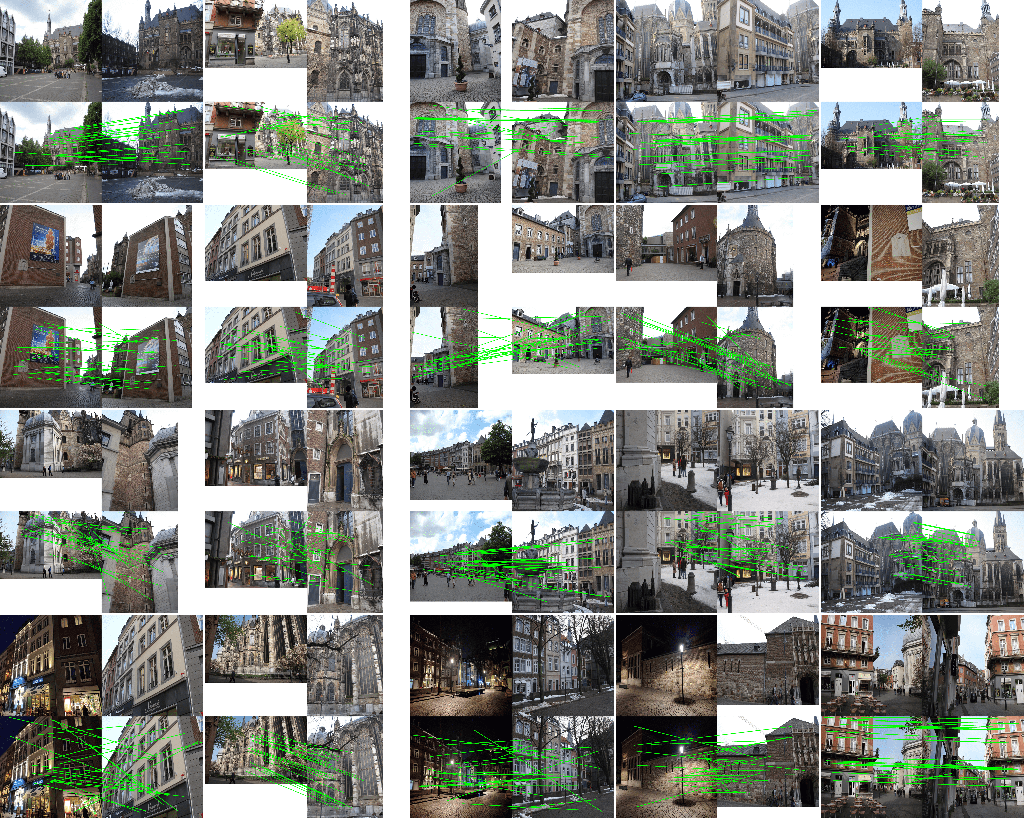}
    \caption{Random sample matches found on the Aachen Day-Night test set. We show the original image pairs top and the pairs with a random subset of located correspondences overlaid below.}
    \label{fig:aachen_supp}
\end{figure*}

\begin{figure*}
    \centering
    \includegraphics[width=\linewidth]{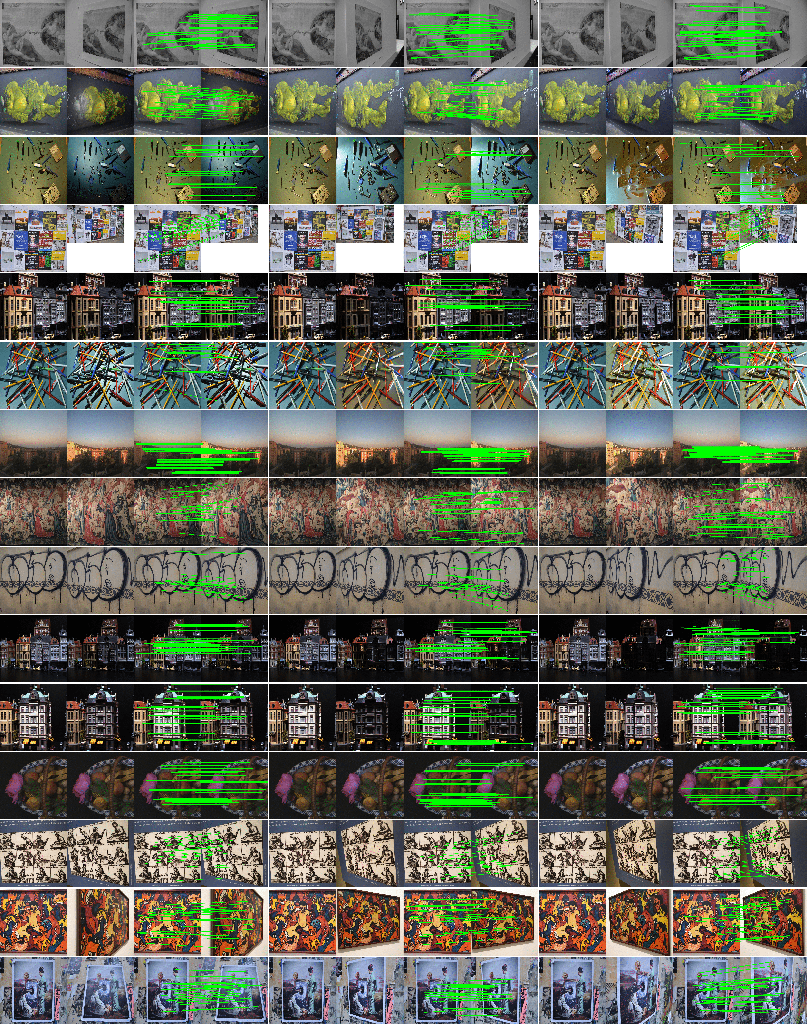}
    \caption{Random sample matches found on the HPatches test set. Rows show progressively harder illumination or viewpoint changes. We first show the original image pairs followed by the pairs with a random subset of located correspondences overlaid.}
    \label{fig:hpatches_supp}
\end{figure*}

\begin{figure*}
    \centering
    \begin{overpic}[width=0.8\linewidth]{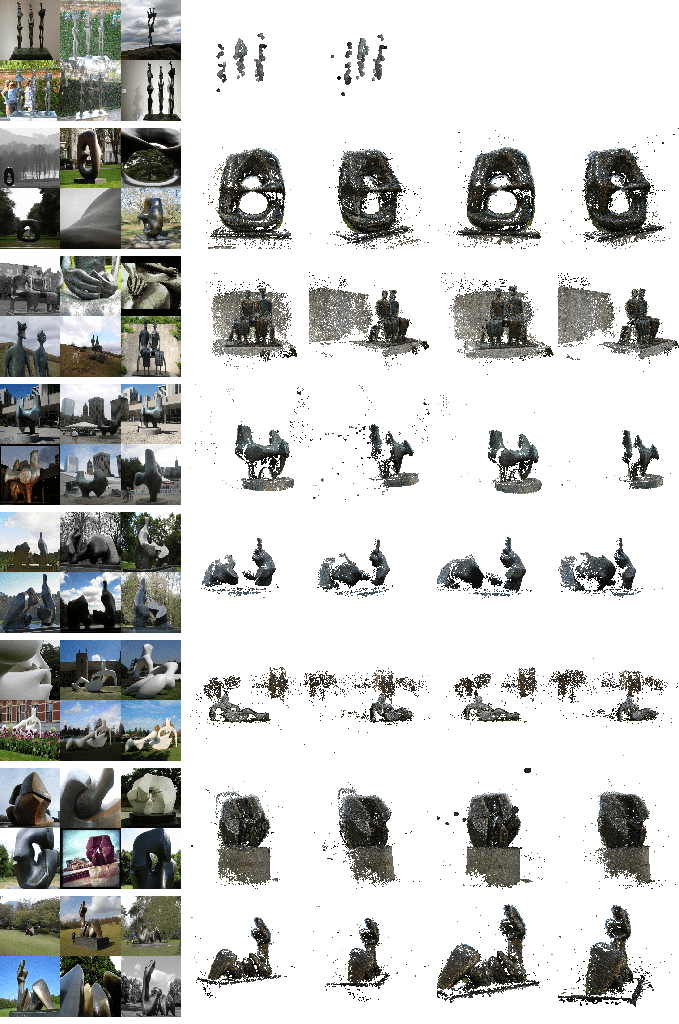}
    \put(5,-1.5){\footnotesize Input Images}
    \put(23,-1.5){\footnotesize Our 3D Reconstruction} 
    \put(42,-1.5){\footnotesize SIFT \cite{Lowe04} 3D Reconstruction }
    \end{overpic}
    \vspace{1em}
    \caption{{\bf Additional SfM results.} Randomly selected input images and 3D models reconstructed using our matches and those obtained using the SIFT \cite{Lowe04} baseline. This figure demonstrates the variety of the input images and their {\em scene shift} as well as that both our and \cite{Lowe04} are of similar quality for these image sets. However, unlike \cite{Lowe04}, our model is able to determine that there are 3 statues in the first example. }
    \label{fig:sfm_supp}
\end{figure*}

\begin{figure*}
    \centering
    \begin{overpic}[height=0.95\textheight]{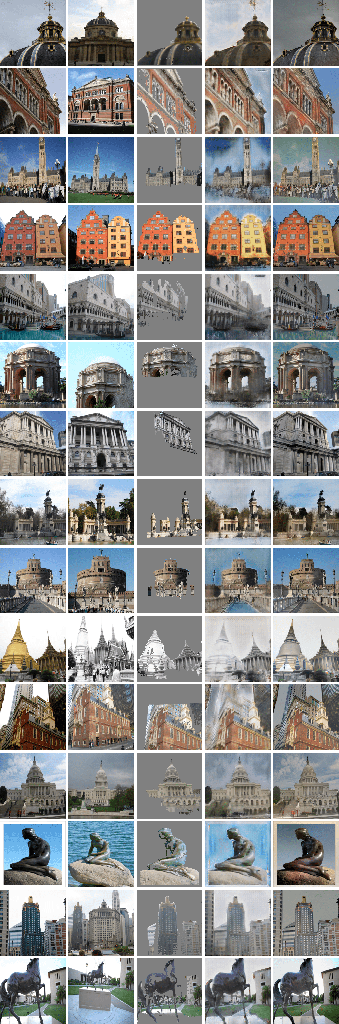}
        \put(3,-1.5){\footnotesize $I_v$}
        \put(9,-1.5){\footnotesize $I_s$}
        \put(15,-1.5){\footnotesize GT$_S$}
        \put(22,-1.5){\footnotesize Ours}
        \put(29,-1.5){\footnotesize \cite{Luan17}}
    \end{overpic}
        \vspace{2em}
    \caption{Additional stylization results. These results are random sampled from the test set of MegaDepth.}
    \label{fig:stylise_supp}
\end{figure*}
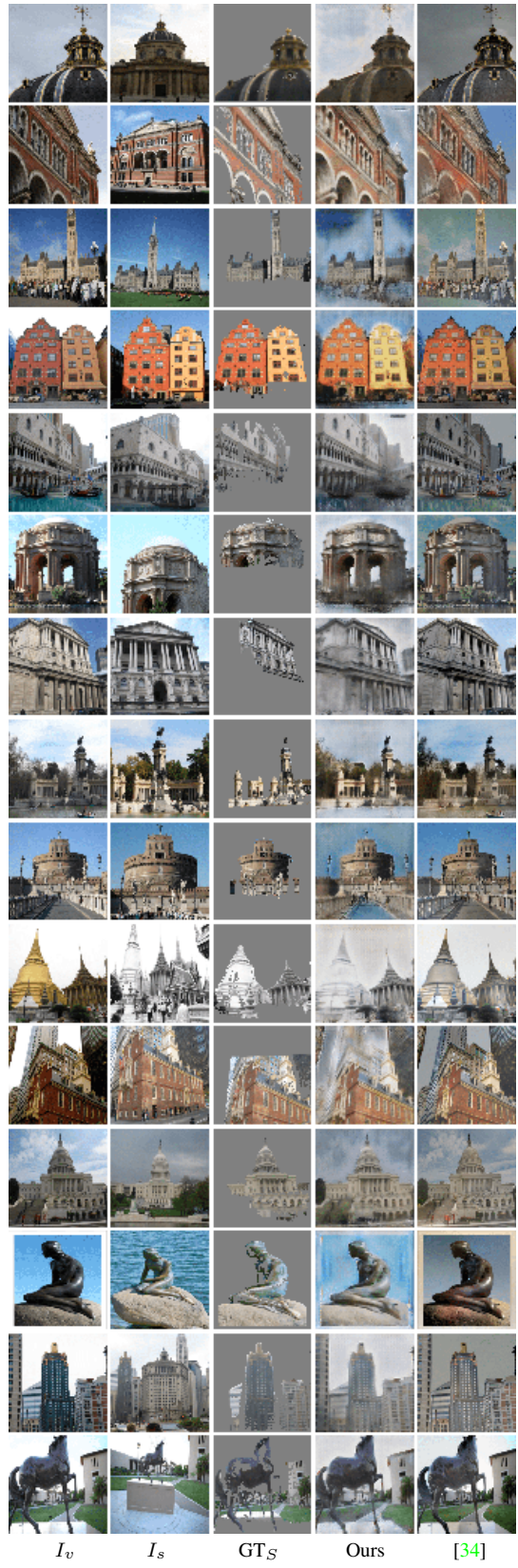